\documentclass[11pt]{article}

\usepackage[final]{acl}

\usepackage{times}
\usepackage{latexsym}
\usepackage[T1]{fontenc}
\usepackage[utf8]{inputenc}
\usepackage{microtype}
\usepackage{inconsolata}

\usepackage{graphicx}

\usepackage{booktabs}
\usepackage{multirow}
\usepackage{array}
\usepackage{tabularx}
\usepackage{calc}

\usepackage{amsmath}
\usepackage{amsfonts}
\usepackage{amssymb}

\usepackage{algorithm}
\usepackage{algpseudocode}

\usepackage{enumitem}
\usepackage{comment}
\usepackage{listings}
\usepackage{xcolor}
\usepackage{wrapfig}
\usepackage{ragged2e}
\usepackage{caption}
\usepackage{subcaption}
\usepackage{pgfplots}
\pgfplotsset{compat=1.18}
\usepackage{tikz}
\usetikzlibrary{shapes.geometric, arrows.meta, positioning, calc, backgrounds, decorations.pathreplacing, fit}
\usepackage{cuted}
\usepackage{caption}   % already loaded by acl.sty, but harmless

\definecolor{darkgreen}{rgb}{0.0, 0.5, 0.0}
\definecolor{firebrick}{rgb}{0.7, 0.13, 0.13}

\lstdefinelanguage{json}{
  basicstyle=\small\ttfamily,
  numbers=left,
  numberstyle=\tiny,
  stepnumber=1,
  numbersep=8pt,
  showstringspaces=false,
  breaklines=true,
  frame=lines,
  string=[s]{"}{"},
  stringstyle=\color{blue!80!black},
  comment=[l]{:},
  commentstyle=\color{black},
}

\title{From Cause to Effect: Generating Intermediate Event Chains in Videos}

\author{
  Ridouane Ghermi$^1$, Vicky Kalogeiton$^1$, Ivan Laptev$^2$ \\
  $^1$LIX, Ecole Polytechnique, IP Paris $^2$MBZUAI \\
  \url{https://ridouaneg.github.io/ccg}
}
\begin{document}
\maketitle

\begin{strip}
    \centering
    \resizebox{\textwidth}{!}{% Teaser figure for Causal Chain Generation (CCG).
% Reuses the cropped frames in images/teaser/ (f1..f6).
\begin{tikzpicture}[
  font=\scriptsize,
  frm/.style={inner sep=0pt, draw=black!55, line width=0.4pt},
  ev/.style={rounded corners=2pt, line width=0.5pt, text width=2.45cm,
             align=center, inner sep=2.4pt, minimum height=1.0cm, anchor=north},
  anc/.style={ev, fill=blue!7, draw=blue!55},
  gen/.style={ev, fill=orange!12, draw=orange!65},
  flow/.style={-{Latex[length=1.6mm,width=1.4mm]}, draw=black!60, line width=0.6pt},
  ts/.style={font=\tiny\sffamily, text=black!60, inner sep=1pt},
  tag/.style={font=\tiny\bfseries, text=blue!60!black},
]
\def\fw{2.55cm}
\def\dx{2.78}

% ---- video frames row ----
\foreach \i/\name/\t in {0/f1/0:46, 1/f2/0:48, 2/f3/0:50, 3/f4/0:52, 4/f5/0:54, 5/f6/1:08}{
  \node[frm] (F\i) at (\i*\dx, 0) {\includegraphics[width=\fw]{images/teaser/\name}};
  \node[ts, above=0pt of F\i] {[\t]};
}
\node[ts, text=black!75, left=2pt of F0, rotate=90, anchor=south, font=\tiny\sffamily\bfseries] {Input video};

% ---- causal chain (event boxes under the frames) ----
\node[anc] (E0) at (0*\dx,-0.68) {A couple takes a photo at a cliff edge};
\node[gen] (E1) at (1*\dx,-0.68) {The photo flies out of the camera};
\node[gen] (E2) at (2*\dx,-0.68) {The woman reaches out to catch it};
\node[gen] (E3) at (3*\dx,-0.68) {Her hiking stick slips off her backpack};
\node[gen] (E4) at (4*\dx,-0.68) {The stick falls off the cliff};
\node[anc] (E5) at (5*\dx,-0.68) {The stick hits a groundhog below};
\foreach \a/\b in {0/1,1/2,2/3,3/4,4/5}{\draw[flow] (E\a.east) -- (E\b.west);}

% start / end tags below the chain + brace under the generated events
\node[tag, below=2pt of E0] {Start event (given)};
\node[tag, below=2pt of E5] {End event (given)};
\draw[decorate, decoration={brace, amplitude=4pt, mirror, raise=1pt}, draw=orange!70!black, line width=0.pt]
  (E1.south west) -- (E4.south east)
  node[midway, below=5pt, font=\scriptsize\bfseries, text=orange!55!black] (brlab) {Intermediate causal chain generated by the model};

% ---- Causal-QA panel ----
\node[draw=darkgreen!55, rounded corners=3pt, fill=darkgreen!4, line width=0.6pt,
      text width=16.0cm, inner sep=5pt, align=left, below=6pt of brlab] (qa) {%
  \textbf{Causal-QA metric.}\quad\textbf{Question: Why does the hiking stick fall from the cliff?}\\[2pt]
  \begin{tabular}{@{}p{7.6cm}@{\hspace{4mm}}p{7.6cm}@{}}
    A.~The woman drops it on purpose. &
    \textcolor{darkgreen}{\textbf{B.~It slips from her backpack as she reaches for the flying photo.}}~\textcolor{darkgreen}{\checkmark}\\
    C.~A groundhog knocks it off the edge. &
    D.~It is blown off the cliff by the wind.\\
    E.~Cannot be determined from the chain. & \\
  \end{tabular}%
};

\end{tikzpicture}
}
    
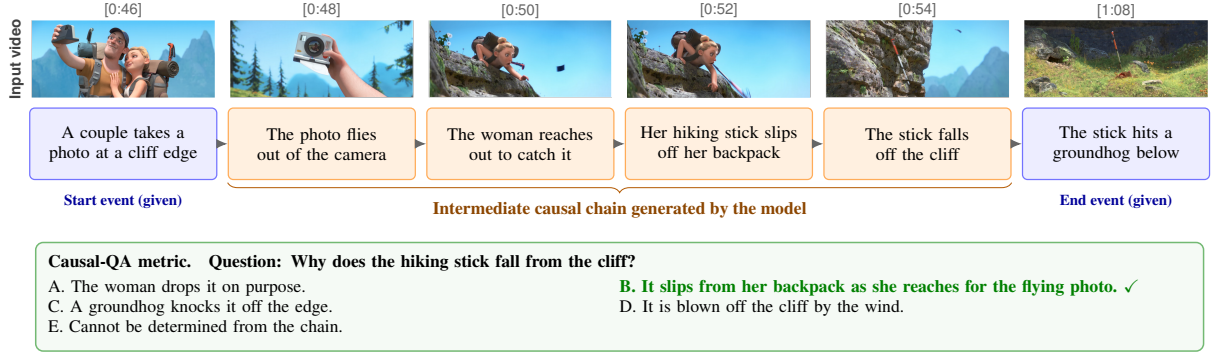
\captionof{figure}{\textbf{Causal Chain Generation (CCG).} Given a video together with a \textcolor{blue!60!black}{start} and \textcolor{blue!60!black}{end} event, the model must generate the \textcolor{orange!60!black}{intermediate causal chain} that bridges them (here: how a selfie leads to a groundhog being hit by a hiking stick). We score the chain with \textbf{Causal-QA}, which asks a fixed reader LLM multiple-choice questions answerable only from the intermediate events (and with SODA-Embedding, not shown).}
    \label{fig:main_figure}
\end{strip}

\begin{abstract}
Video-Language Models (VLMs) excel at visual recognition but struggle with causal reasoning. Current evaluations are flawed: multiple-choice QA allows shortcuts, and open-ended QA lacks robust automatic evaluation. We introduce \textbf{Causal Chain Generation (CCG)}, a novel task where models have to generate the sequence of intermediate events connecting a cause and an effect. This structured setup demands genuine reasoning and enables reliable automatic scoring. We accompany this task with two complementary metrics: \textbf{SODA-Embedding}, which measures the semantic matching of a generated chain against a reference, and \textbf{Causal-QA}, which probes the chain's causal coherence by asking whether it contains the specific causal facts needed to reconstruct the narrative. We provide two test sets: \textbf{SF20K-CC}, a manually annotated diagnostic on 20 short films, and \textbf{CMD-CC}, a movie clip benchmark derived from audio descriptions. Our analysis reveals that state-of-the-art VLMs struggle with narrative causality understanding, frequently hallucinating or omitting key events. To address this, we fine-tune VLMs with GRPO using a composite reward that combines a format reward, SODA-Embedding, and Causal-QA accuracy, demonstrating substantial improvements over strong baselines.
\end{abstract}

\section{Introduction}
\label{sec:introduction}

Understanding \textit{what} happens in a video is only part of the story. True video comprehension requires reasoning about \textit{why} events unfold the way they do, tracing the causal mechanisms that connect an inciting incident to its outcome. This form of narrative causality is central to how humans understand stories, plan actions, and learn from observation. Yet, despite the remarkable progress of recent Video-Language Models (VLMs)~\cite{gemini25, qwen3vl, videosalmonn2, llavavideo} on tasks ranging from video question answering~\cite{videomme, minerva, hourvideo, lvbench, egoschema, temp_compass, mv_bench} to dense video captioning~\cite{vidchapters, activitynet_captions} and object tracking~\cite{perception_test}, their ability to reason about causality remains largely untested, in part because we lack the right tools to measure it.

The core difficulty lies in evaluation. The two dominant paradigms each fall short for causal reasoning. Multiple-choice QA (MCQA)~\cite{nextqa, star, minerva, vited, wtd, mecd, mecd_plus} is easy to score but easy to game: with only a handful of options, models exploit superficial cues rather than reason~\cite{causalvidqa}. Dense Video Captioning (DVC)~\cite{vidchapters, activitynet_captions} requires open-ended generation, but its metrics fall short: temporal-localization metrics (tIoU, F1) ignore semantic content, while n-gram metrics (CIDEr, METEOR, SODA) capture surface word overlap and miss whether the narrative is causally coherent. We thus lack a benchmark that is both \textit{complex enough} to require genuine causal reasoning and \textit{structured enough} to score automatically.

To address this, we introduce a new task: \textbf{Causal Chain Generation (CCG)}. As illustrated in Figure~\ref{fig:main_figure}, given a video, a start event, and an end event, the model must generate the sequence of intermediate events that causally bridges the two. By requiring open-ended generation of a structured sequence, CCG avoids the shortcut problem of MCQA while enabling automatic evaluation, something that free-form open-ended QA cannot reliably achieve. We accompany this task with two complementary evaluation metrics. \textbf{SODA-Embedding} extends the SODA metric~\cite{soda} by replacing n-gram matching with semantic embeddings, shifting the focus from temporal localization to semantic alignment. \textbf{Causal-QA} is a question-answering-based metric: for each chain, we automatically generate a bank of multiple-choice questions whose answers are determinable only from the intermediate events, and at evaluation time we measure the accuracy of a fixed reader LLM that consumes only the model's generated chain. Together, these metrics capture both \emph{local semantic matching} and \emph{global causal coherence}. We provide two complementary datasets: \textbf{SF20K-CC}, a manually annotated diagnostic benchmark on 20 amateur films from SF20K, and \textbf{CMD-CC}, a larger-scale dataset of movie clips from Condensed Movies with annotations derived from audio description (AD), including train and test splits.

Our analysis (Section~\ref{subsec:baselines}) reveals that state-of-the-art VLMs struggle with causal chain generation: common failure modes include missing critical events, hallucinating non-existent ones, and producing causally incoherent sequences. To establish a strong baseline for future work, we benchmark a recipe combining supervised fine-tuning and GRPO with a composite reward derived directly from our two metrics (Section~\ref{sec:method}), and ablate each component.

This paper makes three contributions. We introduce Causal Chain Generation (CCG), a new task for evaluating narrative causality in videos, and accompany it with two complementary metrics, SODA-Embedding and Causal-QA, and two datasets, SF20K-CC (manually annotated full silent films) and CMD-CC (a larger-scale, AD-derived set of movie clips). We then provide a detailed analysis of state-of-the-art VLMs on the task, highlighting systematic failure modes in causal reasoning. Finally, we establish a strong baseline recipe that combines SFT and GRPO with a composite verifiable reward, ablating the contribution of each reward term to give a reference point for future work on video-based causal reasoning.

\begin{comment}
\begin{table}[tb]
  \caption{
  \textbf{Dataset statistics.} 
  %SF20K-CC is manually annotated; CMD-CC is automatically extracted from audio descriptions on a disjoint set of movies.
  }
  \label{tab:benchmark_stats}
  \centering
  \small
  \setlength{\tabcolsep}{3pt}
  \resizebox{\columnwidth}{!}{%
  \begin{tabular}{@{}lrrrrr@{}}
    \toprule
    Test set    & \#Vid & \#Chains & Mean len & \#MCQs & Chain dur. (s) \\
    \midrule
    SF20K-CC-Test & 20    & 205    & 7.7 & 498    & 94.1 \\
    CMD-CC-Test   & 189   & 577    & 7.2 & 1,254  & 46.9 \\
    CMD-CC-Train  & 6,497 & 14,175 & 6.1 & 26,726 & 47.6 \\
  \bottomrule
  \end{tabular}
  }
\end{table}
\end{comment}

\begin{table}[tb]
  \caption{
  \textbf{Dataset statistics.} \#Vid counts distinct source videos (full short films for SF20K-CC, 2-minute movie clips for CMD-CC), \#Mov counts distinct movies, Mean len. is the average number of events per chain, and chain duration is in seconds.
  }
  \label{tab:benchmark_stats}
  \centering
  \small
  \setlength{\tabcolsep}{3pt}
  \resizebox{\columnwidth}{!}{%
  \begin{tabular}{@{}llrrrrrr@{}}
    \toprule
    Dataset  & Split  & \#Vid & \#Mov & \#Chains & Mean len. & \#MCQs & Chain dur. (s) \\
    \midrule
    SF20K-CC & Test   & 20    & 20    & 205       & 7.67 & 537    & 96.9 \\
    CMD-CC   & Test   & 471   & 98    & 1,031    & 6.17 & 2,001  & 69.1 \\
    CMD-CC   & Train  & 6,497 & 1,320 & 14,175   & 6.12 & 26,726 & 68.6 \\
  \bottomrule
  \end{tabular}
  }
\end{table}

\section{Related Work}
\label{sec:related_work}

\noindent\textbf{Video reasoning benchmarks.}
Video understanding benchmarks have evolved from low-level perception such as action recognition~\cite{kinetics, activitynet} toward high-level reasoning. Standard multiple-choice benchmarks~\cite{nextqa, star, egoschema, mv_bench, temp_compass} drove much of this progress but remain prone to shortcut learning, where models exploit answer-option biases rather than reason over the visual content~\cite{causalvidqa}. This has motivated work that evaluates the reasoning \textit{process} itself, from synthetic counterfactual environments~\cite{clevrer, comphy} to rationale-based assessments~\cite{minerva} and event-rich narrative benchmarks~\cite{causalchaos, funqa}. Some approaches structure causal reasoning explicitly, such as VITED~\cite{vited} with evidence chains and MECD~\cite{mecd, mecd_plus} with video Granger causality, but they treat reasoning as an intermediate step toward a multiple-choice answer. Our task instead makes the causal chain the primary output, which also lets us diagnose the ``thinking drift'' failure mode of ~\cite{wtd}, where generated reasoning decouples from the visual evidence.

\noindent\textbf{Dense video captioning.}
Dense Video Captioning (DVC)~\cite{activitynet_captions, vidchapters} is the closest task to ours, as both generate a sequence of event descriptions over a video, but the two differ in fundamental ways. DVC aims for \textit{completeness}, describing all salient events, whereas CCG requires \textit{selectivity}, keeping only events causally relevant to a narrative arc. DVC descriptions are typically context-agnostic (``a man picks up an object''), whereas our events are context-dependent (``the detective retrieves the evidence''). DVC is also evaluated by temporal grounding (tIoU) and surface text overlap (CIDEr, METEOR), whereas SODA-Embedding rewards correct reasoning regardless of phrasing and Causal-QA probes causal coherence through a fixed-judge protocol. Audio descriptions~\cite{cmdad} are a specialized form of dense captioning for accessibility, capturing narrative-relevant content and serving as a useful training source (Section~\ref{sec:method}), though constrained to dialogue gaps and focused on character identification and atmosphere rather than causal reasoning.

\noindent\textbf{VLM post-training.}
Post-training has become central to eliciting reasoning in (vision-)language models. Supervised fine-tuning on curated instruction data is a standard first stage~\cite{liu2023visual, lin2023videollava, liu2024llavanext}, and reinforcement learning, in particular Group Relative Policy Optimization (GRPO)~\cite{deepseekmath, deepseekr1}, has shown that reasoning can be amplified without large-scale supervision~\cite{rlhf}. Such methods are only beginning to reach video~\cite{video-r1}, where designing robust, verifiable rewards remains open. We adopt GRPO and contribute task-specific verifiable rewards, using SODA-Embedding as a dense semantic signal and Causal-QA accuracy as a complementary, more directly causal one.

\section{Causal Chain Generation}
\label{sec:task}

We now formally introduce Causal Chain Generation (CCG). We first define the task (Section~\ref{subsec:task_definition}), then present two complementary evaluation metrics (Section~\ref{subsec:metric_definition}), and finally describe the construction of the two datasets we release with the task (Section~\ref{subsec:benchmark_creation}).

\subsection{Task definition}
\label{subsec:task_definition}

Given a video $\mathcal{V}$ and a query $\mathcal{Q}$ specifying a start event $E_{\text{start}}$ and an end event $E_{\text{end}}$, the goal of the task is to generate the sequence of intermediate events $\mathcal{C} = (E_1, \dots, E_n)$ that causally bridges the two. Each event $E_i$ is a natural language sentence describing a specific, atomic action or state change. We illustrate this formulation in Figure~\ref{fig:main_figure}.

%The task rests on three core design principles. Minimality requires the smallest set of causally sufficient events, granularity ensures that each event represents a single action, and context-dependence means descriptions reflect narrative roles rather than generic actions. Because multiple valid explanations may exist, our evaluation metrics (Section~\ref{subsec:metric_definition}) prioritize semantic similarity and causal coherence over exact textual matches.

The task rests on three core design principles. \emph{Minimality} requires the
smallest set of causally sufficient events, \emph{granularity} ensures that
each event represents a single action, and \emph{context-dependence} means
descriptions reflect narrative roles rather than generic actions. Sections
\ref{subsubsec:sodae} and \ref{subsubsec:causalqa} describe how each
principle is reflected in the metrics. Because multiple valid explanations may
exist, our evaluation metrics prioritize semantic similarity and causal
coherence over exact textual matches.

% Four narrative causality types, one causal chain each, in a 2x2 grid.
% Frames cropped from the original causality_types.png into images/ctypes/.
\newcommand{\ctpanel}[6]{%
% #1 type name | #2 image prefix | #3-#6 frame captions
\begin{tikzpicture}[
  font=\tiny,
  frm/.style={inner sep=0pt, draw=black!45, line width=0.3pt},
  capt/.style={text width=2.0cm, align=center, font=\tiny, inner sep=1pt, anchor=north, text=black!80},
  ar/.style={-{Latex[length=1.1mm,width=0.9mm]}, draw=black!55, line width=0.4pt},
]
  \node[frm] (p1) {\includegraphics[width=2.0cm]{images/ctypes/#21}};
  \node[frm, right=1.8mm of p1] (p2) {\includegraphics[width=2.0cm]{images/ctypes/#22}};
  \node[frm, right=1.8mm of p2] (p3) {\includegraphics[width=2.0cm]{images/ctypes/#23}};
  \node[frm, right=1.8mm of p3] (p4) {\includegraphics[width=2.0cm]{images/ctypes/#24}};
  \foreach \a/\b in {p1/p2,p2/p3,p3/p4}{\draw[ar] (\a.east) -- (\b.west);}
  \node[capt] at (p1.south) {#3};
  \node[capt] at (p2.south) {#4};
  \node[capt] at (p3.south) {#5};
  \node[capt] at (p4.south) {#6};
  \node[anchor=south west, font=\footnotesize\bfseries, text=black!85] at ([yshift=0.3mm]p1.north west) {#1};
\end{tikzpicture}}

\begin{figure*}[t]
  \centering
  \setlength{\tabcolsep}{6pt}
  \renewcommand{\arraystretch}{1.25}
  \resizebox{\textwidth}{!}{%
  \begin{tabular}{c|c}
    \resizebox{0.45\textwidth}{!}{\ctpanel{Physical}{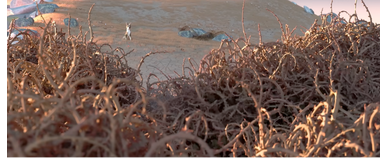}{A dog pushes a sheep into the full house}{A sheep is thrown out of the window}{The sheep rolls down the hill}{It ends up in the brambles}} &
    \resizebox{0.45\textwidth}{!}{\ctpanel{Psychological}{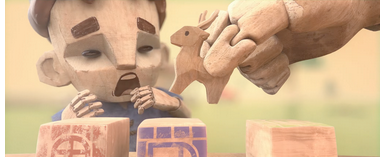}{A father gifts a wooden toy to his son}{The boy ignores his father}{The father puts the toy on the boy's cubes}{The boy screams in fear}} \\ \hline
    \resizebox{0.45\textwidth}{!}{\ctpanel{Motivational}{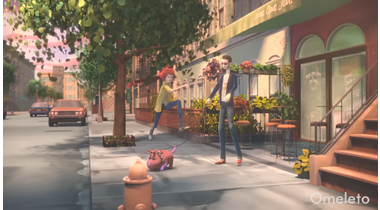}{A man plays guitar in the park}{The chords spread magic glitter}{The glitter reaches a woman}{She runs to reach the park}} &
    \resizebox{0.45\textwidth}{!}{\ctpanel{Enabling}{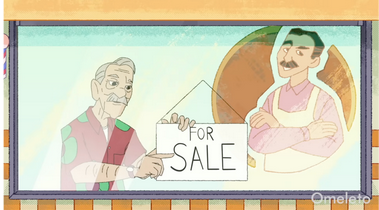}{The old hairdresser puts his salon on sale}{He points to the tip jar}{The young hairdresser sees it full of money}{He buys the hair salon}} \\
  \end{tabular}}
  \caption{\textbf{Narrative causality types.} One representative causal chain for each of the four types in our benchmark, Physical, Psychological, Motivational, and Enabling, read left to right.}
  \label{fig:example}
\end{figure*}

\begin{figure}[t]
\centering
\includegraphics[width=0.9\linewidth]{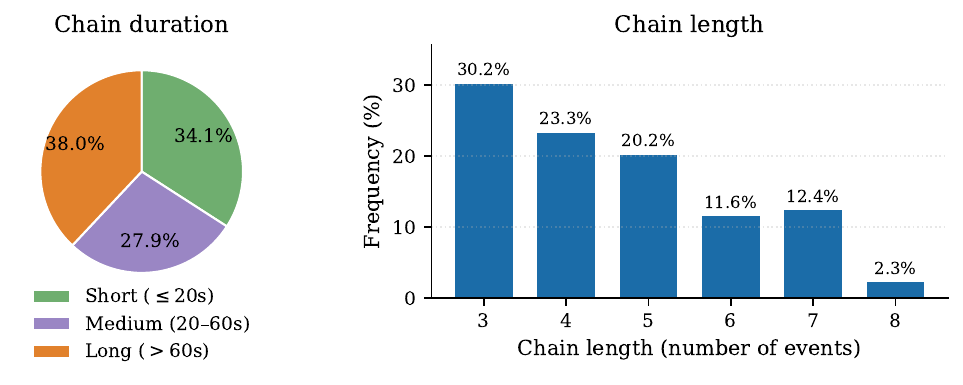}
\caption{\textbf{Causal chain statistics.} \textit{Left:} Distribution of chain durations (in seconds). \textit{Right:} Frequency distribution of chain lengths (in number of events).}
\label{fig:duration_distr}
\end{figure}

\subsection{Evaluation metrics}
\label{subsec:metric_definition}

We evaluate generated causal chains along two complementary axes: \emph{semantic matching}, captured by SODA-Embedding (Section~\ref{subsubsec:sodae}), and \emph{causal coherence}, captured by Causal-QA (Section~\ref{subsubsec:causalqa}). The two metrics target different failure modes: SODA-Embedding rewards correct phrasing and ordering of individual events, while Causal-QA asks whether the chain contains the specific causal details needed to reconstruct the narrative.

\subsubsection{SODA-Embedding}
\label{subsubsec:sodae}

SODA-Embedding measures the semantic and sequential quality of a generated chain against a ground-truth reference. It extends the SODA framework~\cite{soda}, originally designed for dense video captioning, with two key modifications. We discard the temporal Intersection-over-Union (tIoU) constraint, since our task prioritizes the logical ordering of events over precise temporal grounding, and align events purely by their semantic and sequential correspondence. We also replace the n-gram matching score (METEOR) with cosine similarity between text embeddings (Qwen3-Embedding), which rewards semantically correct reasoning even when phrased differently from the reference.

Concretely, given a predicted chain $P = (p_1, \dots, p_m)$ and a ground-truth chain $G = (g_1, \dots, g_n)$, we compute a pairwise cosine similarity matrix between all event embeddings. We then find the optimal order-preserving alignment through dynamic programming, similar to Dynamic Time Warping, yielding a soft true-positive count. From this, we derive precision (normalized by $m$), recall (normalized by $n$), and report their harmonic mean as the final SODA-Embedding score (full algorithm in the Appendix, with an example in Figure~\ref{fig:metric_illustration}). This makes the metric robust to paraphrasing (via embeddings), strict on ordering errors (via the sequential alignment), and length-agnostic (via the F1 formulation).

\subsubsection{Causal-QA}
\label{subsubsec:causalqa}

While SODA-Embedding captures local semantic alignment, it does not directly test whether the generated chain contains the \emph{specific causal facts} needed to explain the narrative. A chain can be lexically and ordering-wise close to the reference yet still miss or distort the critical causal link. To probe this directly, we introduce Causal-QA, a question-answering–based evaluation.

\noindent\textbf{Definition.}
For each ground-truth chain, we associate a small bank of five-way multiple-choice questions (MCQs) whose answers are determinable \emph{only} from the chain's intermediate events. At evaluation time, we provide the model's generated chain as the sole context to a fixed reader LLM and measure its accuracy on the bank. A chain that omits or hallucinates the key causal events will leave the reader unable to recover the answer, resulting in a low score. We show example items in Figure~\ref{fig:main_figure} and Appendix~\ref{sec:causalqa_examples} (Figure~\ref{fig:causalqa}).

\noindent\textbf{Generation.}
For each ground-truth chain, a generator LLM (Qwen3.6-27B) proposes 6--10 candidate five-option MCQs (A--E), where option E is always ``Cannot be determined from the chain'' and the four distractors are drawn from a fixed set of confusion types (e.g., wrong agent, wrong instrument, swapped order, inverted polarity).

\noindent\textbf{Filtering.}
A fixed judge LLM (Qwen2.5-3B-Instruct) then keeps only questions that pass three sequential tests: \textit{answerable from the full chain} (the judge selects the gold answer given the full chain, removing ill-posed questions), \textit{unanswerable with no context} (the judge does not select it given no context, removing questions that can be guesses from prior knowledge), and \textit{unanswerable from anchors alone} (the judge does not select it given only $E_{\text{start}}$ and $E_{\text{end}}$, so the question only probes the intermediate chain). Surviving questions form the Causal-QA bank. We perform majority voting over multiple predictions to avoid dropping questions that are answered by chance.
To verify that the surviving questions are well-posed, two annotators answered 150 sampled Causal-QA questions given the full ground-truth chain, i.e.\ the setting in which the metric operates. Human accuracy reaches 94.2\% on average (five-way MCQ, chance 20\%) with strong inter-annotator agreement (Fleiss' $\kappa = 0.853$), confirming that the questions are reliably answerable from the chain alone
%(details in Appendix~\ref{supmat:causalqa_human})
.

\noindent\textbf{Evaluation setup.}
At evaluation time, we score a generated chain by passing it to the same fixed judge LLM and reporting accuracy across the clip's Causal-QA MCQs. Crucially, both the judge and the prompt are fixed and disclosed: this gives a calibrated, deterministic reader. This is in contrast to LLM-as-a-judge approaches that are sensitive to model scale and prompt choice~\cite{chen2024humansllmsjudgestudy, zheng2023judgingllmasajudgemtbenchchatbot, panickssery2024llmevaluatorsrecognizefavor}. We use Qwen2.5-3B-Instruct as the fixed judge, decoded greedily, and perform robustness analysis with different model types and sizes (see Section~\ref{subsec:metric_validation}).

\subsection{Dataset construction}
\label{subsec:benchmark_creation}

\noindent\textbf{Data sources.}
We construct two complementary datasets that trade scale against narrative completeness. SF20K-CC is a manually annotated diagnostic set of 20 full silent animation films from SF20K~\cite{sf20k}. Being amateur, complete, and dialogue-free, these films limit data contamination, expose long-range causal structure, and remove the dialogue shortcut, at the cost of expensive manual annotation. CMD-CC is a larger dataset built on CMD-AD~\cite{cmdad}, providing a sizable pool with reliable chains derived from manual audio descriptions, at the cost of short clips with truncated narrative arcs. CMD-CC is split into a train set (used in Section~\ref{sec:method}) and a test set drawn from disjoint movies, both produced by the same automated pipeline (below), with the test split additionally manually curated to improve annotation quality.

\noindent\textbf{Annotation protocol.}
Both datasets are built with the same three-step protocol. We first obtain dense audio descriptions (ADs) for every video, writing them manually for the SF20K-CC silent short films and reusing the existing manual ADs of CMD-AD~\cite{cmdad} for CMD-CC. 

From these ADs, we prompt a large language model~\cite{gpt5} to build a full causal graph over the video's events; the exact prompt is given in
Figure~\ref{supmat:prompt_data_gen}. We then review this graph and extract causal chains as sub-parts of it, each taking a start event $E_{\text{start}}$ and an end event $E_{\text{end}}$ together with the intermediate events that causally connect them. For each extracted chain we build the associated Causal-QA bank with the filtering pipeline of Section~\ref{subsubsec:causalqa}, and the CMD-CC test split is additionally manually curated to improve annotation quality.

\noindent\textbf{Causality types.}
To ensure coverage of diverse reasoning patterns, we adopt the taxonomy proposed by Hu et al.~\cite{inferring_causality}, which distinguishes four types of narrative causality: \textit{Physical} (Event A physically causes Event B), \textit{Motivational} (Event A provides the goal or reason for Event B), \textit{Psychological} (Event A triggers an emotional state change leading to Event B), and \textit{Enabling} (Event A creates the necessary conditions for Event B to occur). We classify each chain accordingly. Figure~\ref{fig:example} shows one example per type, and Appendix~\ref{supmat:causality_types} reports the distribution across types and a per-type performance analysis.

\noindent\textbf{Statistics.}
Table~\ref{tab:benchmark_stats} summarizes the two datasets. SF20K-CC-Test comprises 20 full silent short films with 205 manually verified causal chains, and CMD-CC-Test provides 1,031 chains with 2,001 Causal-QA MCQs, drawn from movies disjoint from the 14,175-chain training set. Figure~\ref{fig:duration_distr} shows the distributions of chain durations and lengths on SF20K-CC.

\section{A training recipe for CCG}
\label{sec:method}

While current VLMs can identify isolated events, they struggle to assemble them into coherent causal chains. In this section, we describe a training recipe for CCG that we benchmark in Section~\ref{sec:experiments}. The recipe pairs two off-the-shelf ingredients, each tailored to the task: supervised fine-tuning (SFT) on the CMD-CC train split as a simple baseline, and Group Relative Policy Optimization (GRPO) with a composite verifiable reward whose terms correspond directly to our two metrics. We additionally explore test-time scaling in Appendix~\ref{supmat:tts}. The novelty of this section is not in the choice of algorithms but in the design of the reward signal, which we show is the key driver of performance in Section~\ref{subsec:reward_ablation}.

\subsection{Supervised fine-tuning}
\label{subsec:sft}

As a baseline training recipe, we fine-tune the model to maximize the likelihood of the ground-truth causal chain $\mathcal{C}_{\text{gt}}$ via standard cross-entropy (full objective in Appendix~\ref{supmat:training_losses}). SFT is simple and stable, but it penalizes any deviation from the exact phrasing of the reference, even when the generated chain is semantically correct or conveys the same causal coherence in different words. This rigidity is suboptimal for an open-ended task like CCG, motivating the RL-based recipe below.

\subsection{GRPO with a composite verifiable reward}
\label{subsec:finetuning}

\noindent\textbf{Objective.}
We employ Group Relative Policy Optimization (GRPO)~\cite{deepseekmath}, which optimizes the policy $\pi_\theta$ to maximize a task-specific composite reward. For each training query $q = (\mathcal{V}, E_{\text{start}}, E_{\text{end}})$, we sample a group of $G$ candidate chains $\{o_1, \dots, o_G\}$ from the old policy $\pi_{\theta_{\text{old}}}$ and update $\pi_\theta$ with the standard clipped GRPO objective using the group-relative advantage $A_i = (R(o_i) - \mu_R) / \sigma_R$, where each chain's reward $R(o_i)$ is normalized within the group. The full objective is given in Appendix~\ref{supmat:training_losses}.

\noindent\textbf{Composite reward.}
The core design choice of our recipe is the structure of the reward. We use a sum of three complementary, non-differentiable signals, each targeting a distinct failure mode:
\begin{multline}
\label{eq:reward}
    R(o_i) = R_{\text{format}}(o_i) + R_{\text{SODA-E}}(o_i, \mathcal{C}_{\text{gt}}) \\
    + R_{\text{Causal-QA}}(o_i, \mathcal{Q}_{\text{gt}})
\end{multline}

$R_{\text{format}}(o_i)$ is a binary reward that requires the output to be valid JSON containing a list of events. $R_{\text{SODA-E}}(o_i, \mathcal{C}_{\text{gt}}) \in [0, 1]$ is the SODA-Embedding score (Section~\ref{subsubsec:sodae}) between the generated chain $o_i$ and the ground-truth chain $\mathcal{C}_{\text{gt}}$, rescaled from $[0, 100]$, providing a dense signal that captures local semantic matching and correct event ordering. $R_{\text{Causal-QA}}(o_i, \mathcal{Q}_{\text{gt}}) \in [0, 1]$ is the accuracy of the fixed judge LLM on the MCQ bank $\mathcal{Q}_{\text{gt}}$ when conditioned only on $o_i$ (Section~\ref{subsubsec:causalqa}). This is a sparser but more directly causal signal, since a chain that omits or distorts a key intermediate event fails to support the correct MCQ answer regardless of how fluent it is.

%Runs that include $R_{\text{Causal-QA}}$ are trained on the subset of CMD-CC clips that have a Causal-QA bank ($\mathcal{D}_{\text{train}}^{\text{QA-only}}$). The relative contributions of $R_{\text{SODA-E}}$ and $R_{\text{Causal-QA}}$ are studied empirically in Section~\ref{subsec:reward_ablation}.

% Composite GRPO reward: clean horizontal flow (one-column).
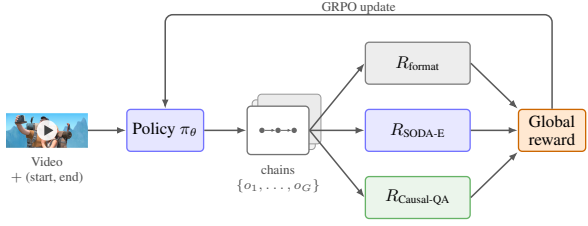
\begin{figure}[t]
\centering
\resizebox{\linewidth}{!}{%
\begin{tikzpicture}[
  font=\small, >={Latex[length=1.7mm]},
  box/.style={rounded corners=2pt, line width=0.7pt, align=center, inner sep=3pt, minimum height=0.78cm},
  pol/.style={box, draw=blue!55, fill=blue!10, minimum width=1.35cm},
  rf/.style={box, draw=black!50, fill=black!7, minimum width=1.95cm},
  rs/.style={box, draw=blue!55, fill=blue!9, minimum width=1.95cm},
  rq/.style={box, draw=darkgreen!60, fill=darkgreen!8, minimum width=1.95cm},
  glob/.style={box, draw=orange!80!black, fill=orange!18, minimum width=1.25cm},
  ar/.style={->, draw=black!65, line width=0.8pt},
  lbl/.style={font=\scriptsize, text=black!70},
]
% ---- input: video + query ----
\node[inner sep=0pt, draw=black!55, line width=0.5pt] (vid) at (0,0)
  {\includegraphics[width=1.5cm]{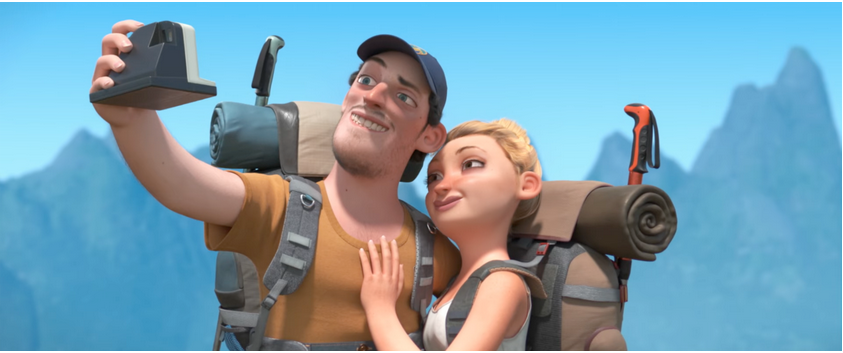}};
\fill[white, opacity=0.9] (vid.center) circle (0.19);
\fill[black!75] (vid.center)++(-0.05,0.1) -- ++(0,-0.2) -- ++(0.17,0.1) -- cycle;
\node[lbl, below=0.4mm of vid, align=center] {Video\\$+$ (start, end)};

% ---- policy ----
\node[pol] (pol) at (2.2,0) {Policy $\pi_\theta$};

% ---- outputs: stack of candidate chains ----
\node[box, draw=black!55, fill=white, minimum width=1.15cm, minimum height=0.9cm] (out) at (4.3,0) {};
\foreach \k in {0,1,2}{\fill[black!60] ($(out.center)+(-0.3+0.3*\k,0)$) circle (0.05);}
\draw[-{Latex[length=0.7mm]}, black!60, line width=0.4pt] ($(out.center)+(-0.25,0)$) -- ($(out.center)+(-0.05,0)$);
\draw[-{Latex[length=0.7mm]}, black!60, line width=0.4pt] ($(out.center)+(0.05,0)$) -- ($(out.center)+(0.25,0)$);
\begin{scope}[on background layer]
  \draw[rounded corners=2pt, draw=black!40, fill=black!5] ([xshift=3pt,yshift=3pt]out.south west) rectangle ([xshift=3pt,yshift=3pt]out.north east);
  \draw[rounded corners=2pt, draw=black!40, fill=black!8] ([xshift=6pt,yshift=6pt]out.south west) rectangle ([xshift=6pt,yshift=6pt]out.north east);
\end{scope}
\node[lbl, below=0.4mm of out, align=center] {chains\\$\{o_1,\dots,o_G\}$};

% ---- rewards ----
\node[rf] (rf) at (6.9,1.25)  {$R_{\text{format}}$};
\node[rs] (rs) at (6.9,0)     {$R_{\text{SODA-E}}$};
\node[rq] (rq) at (6.9,-1.25) {$R_{\text{Causal-QA}}$};

% ---- global reward ----
\node[glob] (g) at (9.4,0) {Global\\reward};

% ---- arrows ----
\draw[ar] (vid) -- (pol);
\draw[ar] (pol) -- (out);
\draw[ar] (out.east) -- (rf.west);
\draw[ar] (out.east) -- (rs.west);
\draw[ar] (out.east) -- (rq.west);
\draw[ar] (rf.east) -- (g.north west);
\draw[ar] (rs.east) -- (g.west);
\draw[ar] (rq.east) -- (g.south west);

% ---- loop back: clean route over the top ----
\draw[ar, rounded corners=4pt] (g.north) -- ++(0,1.75) -| (pol.north);
\node[lbl] at (5.8,2.28) {GRPO update};
\end{tikzpicture}}
\caption{\textbf{Our composite GRPO reward.} The policy samples a group of candidate chains for each query; every chain is scored by three complementary signals, a JSON format reward, the SODA-Embedding alignment to the ground-truth chain, and the Causal-QA accuracy of a fixed reader LLM, which combine into a global reward that drives the GRPO update (Section~\ref{subsec:finetuning}).}
\label{fig:method}
\end{figure}

\section{Experiments}
\label{sec:experiments}

In this section, we first validate our SODA-Embedding metric via synthetic perturbations and human correlation (Section~\ref{subsec:metric_validation}). We then benchmark state-of-the-art VLMs on our task (Section~\ref{subsec:baselines}), evaluate the training recipe of Section~\ref{sec:method} (Section~\ref{subsec:method_results}), and ablate the composition of the composite reward (Section~\ref{subsec:reward_ablation}).

\subsection{Metric validation}
\label{subsec:metric_validation}

\begin{figure}[t]
    \centering
    \includegraphics[width=\linewidth]{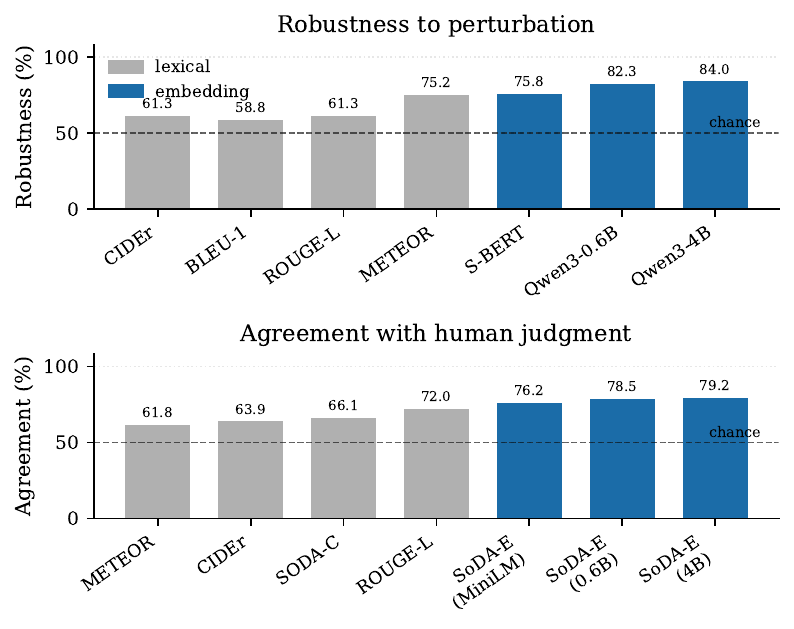}
    \caption{\textbf{SODA-Embedding validation.} \textit{Top:} robustness to perturbation (mean success rate over the four operators, where a success is $\textit{metric}(g, \tilde{p}) < \textit{metric}(g, p)$). \textit{Bottom:} pairwise agreement with human judgment. Embedding-based metrics dominate ngram metrics on both axes; in particular SODA-E beats its ngram ablation SODA-C.}
    \label{fig:metric_validation}
\end{figure}

\noindent\textbf{Perturbation analysis.}
A robust metric should score a high-quality prediction $p$ above a degraded version $\tilde{p}$. We build ($g, p, \tilde{p}$) triplets from a ground-truth $g$, where $\tilde{p}$ applies one of four operators to $p$: \textit{Swapping} (exchange two elements), \textit{Shuffling} (reorder the sequence), \textit{Intruder} (replace one element at random), and \textit{Random} (replace $p$ entirely). A `success' is $metric(g, \tilde{p}) < metric(g, p)$. As Figure~\ref{fig:metric_validation} (top) shows, text embeddings significantly outperform n-gram metrics, and robustness improves with the scale of the base embedding LLM.

\noindent\textbf{Agreement with human judgment.}
We run a pairwise meta-evaluation: an annotator is shown a ground-truth chain $g$ and two predictions ($p_1, p_2$) and picks the better one, and a metric agrees if the prediction it scores higher is the one preferred. Over 256 stratified A/B pairs (128 CMD-CC, 128 SF20K-CC) with three independent annotators in strong agreement (Fleiss $\kappa=0.881$), Figure~\ref{fig:metric_validation} (bottom) reports each metric's pairwise agreement. Embedding-based metrics agree with the annotator far more than ngram ones (SODA-E at 79.2\% versus 61.8--72.0\% for METEOR, CIDEr, and ROUGE-L). Crucially, SODA-E beats its own ngram ablation SODA-C, which has the same alignment method but scores with METEOR (79.2\% versus 66.1\%): the embedding scorer, not the alignment, is what makes the metric better aligned with human judgment. Moreover, SODA-E improves with the embedder quality and size, going from 76.2\% (SODA-E with MiniLM) to 79.2\% (SODA-E with Qwen3-Embedding-4B).

\noindent\textbf{Robustness to the underlying models.}
Each metric relies on an auxiliary model (SODA-Embedding on a text embedder, Causal-QA on a judge LLM), so we verify that our system rankings are not an artifact of those choices by re-scoring all systems while varying only that model. The rankings are remarkably stable, not only across model sizes but, more importantly, across model families. Swapping the SODA-Embedding embedder from Qwen3-Embedding-4B to its 0.6B/8B variants or to the cross-family all-MiniLM-L6-v2 preserves the system ordering ($\ge 91\%$ rank agreement, Spearman $\rho\ge 0.91$). Causal-QA is just as stable when the judge is competent: a 7B Qwen judge and the cross-family Gemma-3-12B and Llama-3.1-8B judges all reproduce the ranking of our default Qwen2.5-3B judge ($\ge 85\%$ agreement). The only exception is a 0.5B judge, too weak to separate systems, which collapses to near-chance agreement on the smaller SF20K set. We therefore fix Qwen3-Embedding-4B and Qwen2.5-3B-Instruct and conclude that our findings are robust to these choices (see Table~\ref{tab:metric_robustness} and Appendix~\ref{supmat:metric_val}).

\subsection{Baselines}
\label{subsec:baselines}

\begin{table*}[t]
\centering
\caption{\textbf{Baseline performance.} Zero-shot performance on CMD-CC and SF20K-CC, grouped by baseline family. The reference conditions isolate each input modality: \emph{Random} shuffles reference chains across test instances, \emph{Guessing} sees the query alone, \emph{Blind} adds subtitles but no frames. \emph{Video}: full video at 1 FPS, or a truncated segment centered on the causal chain (32 frames); the two give comparable scores for the same model (Appendix~\ref{supmat:fullvstrunc}). \emph{Subs.}: subtitle transcript provided. All models receive the ground-truth number of events. Best per column in \textbf{bold}.}
\label{tab:baselines}
\small
\setlength{\tabcolsep}{6.5pt}
\resizebox{2.\columnwidth}{!}{%
\begin{tabular}{@{}lccccrrrr@{}}
\toprule
& & & & & \multicolumn{2}{c}{\textbf{CMD-CC}} & \multicolumn{2}{c}{\textbf{SF20K-CC}} \\
\cmidrule(lr){6-7} \cmidrule(lr){8-9}
Model & Size & Video & Frames & Subs. & SODA-E & CQA & SODA-E & CQA \\
\midrule
\multicolumn{9}{@{}l}{\textit{Reference conditions}} \\
\hspace{1em}Random                & --  & --   & --   & --          & 35.6 & 13.0 & 36.5 & 2.4  \\
\hspace{1em}Guessing (query-only) & --  & --   & --   & $\times$    & 45.6 & 7.5  & 45.8 & 11.6 \\
\hspace{1em}Blind (query + subtitles) & --  & --   & --   & \checkmark  & 52.9 & 20.8 & 45.0 & 11.6 \\
\midrule
\multicolumn{9}{@{}l}{\textit{Open-source VLMs}} \\
\hspace{1em}VideoLLaMA3-7B~\cite{videollama3} & 7B & full & 1 FPS & \checkmark & 34.4 & 9.8  & 36.4 & 8.5  \\
\hspace{1em}InternVL-3.5-8B~\cite{internvl35} & 8B & full & 1 FPS & \checkmark & 53.3 & 18.7 & 53.1 & 13.2 \\
\hspace{1em}GLM-4.6V-Flash~\cite{glmv}        & 9B & full & 1 FPS & \checkmark & 47.2 & 19.1 & 44.6 & 16.7 \\
\hspace{1em}Qwen3.5-9B~\cite{qwen3vl}         & 9B & full & 1 FPS & \checkmark & 54.6 & 21.5 & 52.7 & 16.5 \\
%\hspace{1em}Qwen3.5-27B~\cite{qwen3vl}       & 27B & full & 1 FPS & \checkmark & 59.9 & 24.4 & -- & -- \\
\midrule
\multicolumn{9}{@{}l}{\textit{Causal-specialized baselines}} \\
\hspace{1em}Qwen3.5-9B-Thinking~\cite{qwen3vl} & 9B & full & 1 FPS & \checkmark & 40.3 & 24.9 & 40.0 & 21.7 \\
\hspace{1em}MECD~\cite{mecd}                   & 9B & full & 1 FPS & \checkmark & 44.2 & 10.1 & 48.2 & 12.4 \\
\hspace{1em}TEMPURA-3B~\cite{tempura}          & 3B & full & 1 FPS & \checkmark & 30.5 & 12.0 & 29.6 & 11.3 \\
\midrule
\multicolumn{9}{@{}l}{\textit{Two-stage pipelines}} \\
\hspace{1em}Two-stage (caption $\to$ chain) & 9B & full & 1 FPS & \checkmark & 45.9 & 8.5  & 50.2 & 13.3 \\
\hspace{1em}Two-stage (graph $\to$ chain)   & 9B & full & 1 FPS & \checkmark & 47.4 & 11.2 & 50.1 & 14.3 \\
\midrule
\multicolumn{9}{@{}l}{\textit{Proprietary VLMs}} \\
\hspace{1em}GPT-5.4-nano~\cite{gpt54}              & -- & trunc. & 32 & \checkmark & 52.5 & 22.2 & 52.5 & 15.5 \\
\hspace{1em}GPT-5.4-mini~\cite{gpt54}              & -- & trunc. & 32 & \checkmark & 47.9 & \textbf{29.5} & 52.7 & \textbf{23.4} \\
\hspace{1em}Gemini-3.1-Flash-lite~\cite{gemini31flashlite} & -- & trunc. & 32 & \checkmark & \textbf{61.6} & 26.4 & \textbf{56.0} & 22.6 \\
\bottomrule
\end{tabular}
}
\end{table*}

\noindent\textbf{Setup.}
We benchmark a broad set of systems on SF20K-CC and CMD-CC: open-source VLMs of varying scale, proprietary models, dedicated causal-reasoning baselines, and caption-then-reason pipelines, alongside trivial \textit{Random}, \textit{Guessing}, and \textit{Blind} references (full list, citations, and inference details in Appendix~\ref{supmat:baseline_setup}). Each model is run on either the full video (1 FPS) or a truncated clip centered on the chain (32 frames). These conditions are comparable (Appendix~\ref{supmat:fullvstrunc}), so all rows share one table. All models receive the subtitle transcript, and metrics use the embedder and judge of Section~\ref{subsec:metric_definition}.

\noindent\textbf{Scaling behavior.} Sweeping the Qwen3.5 family (Appendix~\ref{supmat:scaling}) shows performance rising steeply with model size up to 4B and then saturating. Even the strongest baseline (27B, 59.9 SODA-Embedding on CMD-CC) leaves substantial headroom, indicating that CCG is far from solved. Returns in the number of input frames are roughly logarithmic, with Causal-QA the more frame-hungry axis.

\noindent\textbf{Trends.} Model scale is the single largest factor, with a broadly consistent ordering across the two datasets. The chain-of-thought (``thinking'') variant reaches the best Causal-QA among non-finetuned open models (24.9 / 21.7) but at much lower SODA-Embedding, as its verbose reasoning diverges from the reference. The caption-then-reason pipelines underperform the corresponding single-pass VLM on CMD-CC, indicating that decomposing the task loses causally relevant visual detail. GPT-5.4-mini reaches the highest Causal-QA overall (29.5 / 23.4). Since the full-video and truncated-clip conditions are comparable (Appendix~\ref{supmat:fullvstrunc}) and all models receive the ground-truth number of events, this is a genuine gap. Gemini-3.1-Flash-lite, run in the same constrained regime, matches GPT-5.4-mini on SF20K (56.0 SODA-Embedding, 22.6 Causal-QA).

\noindent\textbf{Where difficulty lives.} Slicing CMD-CC Causal-QA along causality type, chain length, and dialogue availability (Appendix~\ref{supmat:breakdown}) shows that dialogue dominates: models score $+11$ to $+20$ Causal-QA points higher on with-dialogue clips than on silent ones, and on the former a text-only control rivals the video model. Stripping the subtitle track from Qwen3.5-9B erases this advantage ($-4.2$ on CMD-CC), confirming it comes from the transcript. The silent subset, where the text-only baseline collapses, is therefore the more challenging evaluation of visual causal reasoning. Chain length, by contrast, is not a primary difficulty axis.

\subsection{Method performance}
\label{subsec:method_results}

\noindent\textbf{Setup.}
We fine-tune the Qwen3.5-9B backbone on our synthetic CMD-AD training chains (Section~\ref{subsec:benchmark_creation}) in two stages, both LoRA ($r=16$, $\alpha=32$, on the q/k/v/o projections) for one epoch each. We first run supervised fine-tuning (SFT) on the reference chains, then apply GRPO with the composite reward of Equation~\ref{eq:reward}: $R_{\text{format}} + R_{\text{SODA-E}} + R_{\text{Causal-QA}}$. Videos are decoded at 1 FPS, matching the baseline setup. All metric values use the same judge and embedder as the baselines.

\begin{table}[t]
    \caption{\textbf{Training-algorithm ablation} of Qwen3.5-9B: no fine-tuning (Base), SFT, GRPO from the base model, and SFT followed by GRPO. All GRPO runs use the full composite reward. Best per column in \textbf{bold}, second-best \underline{underlined}.}
    \label{tab:method_performance}
    \centering
    \small
    \setlength{\tabcolsep}{5pt}
    \begin{tabular}{@{}lrrrr@{}}
    \toprule
    & \multicolumn{2}{c}{\textbf{CMD-CC}} & \multicolumn{2}{c}{\textbf{SF20K-CC}} \\
    \cmidrule(lr){2-3} \cmidrule(lr){4-5}
    Method & SODA-E & CQA & SODA-E & CQA \\
    \midrule
    Base       & 54.6 & 21.5 & 52.7 & 16.5 \\
    SFT        & \underline{62.3} & 24.6 & 56.6 & 19.3 \\
    GRPO       & 58.9 & \textbf{32.2} & \underline{57.2} & \textbf{24.9} \\
    SFT + GRPO & \textbf{63.3} & \underline{29.4} & \textbf{57.9} & \underline{22.1} \\
    \bottomrule
    \end{tabular}
\end{table}

\noindent\textbf{Training algorithm.} Every fine-tuning improves the base model (Table~\ref{tab:method_performance}). SFT gives the largest chain-quality gain ($+7.6$ SODA-Embedding on CMD-CC, to 62.3) but barely moves Causal-QA. GRPO from the base model instead pushes Causal-QA highest (32.2 on CMD-CC, 24.9 on SF20K-CC) at slightly lower SODA-Embedding, since it chases the judge without SFT's fluency prior. Stacking the two (SFT\,+\,GRPO) is best balanced, reaching the top chain quality (63.3 / 57.9 SODA-Embedding) while recovering much of the Causal-QA gain (29.4 on CMD-CC). The two routes thus trade chain quality against causal accuracy, a tension we dissect in the reward ablation below. Because both evaluation metrics are used as rewards during GRPO fine-tuning, we verify these performance gains against a held-out judge: it rises monotonically from the base model through SFT to SFT + GRPO on both test sets (see Table~\ref{tab:heldout_judge} in Appendix~\ref{supmat:heldout_judge}).

\subsection{Reward ablation}
\label{subsec:reward_ablation}

To disentangle the contribution of the two semantic terms in our composite reward (Equation~\ref{eq:reward}), we train Qwen3.5-9B with GRPO directly from the base model (no SFT initialization) under each combination of $R_{\text{SODA-E}}$ and $R_{\text{Causal-QA}}$, with $R_{\text{format}}$ always on. 
Arms that include $R_{\text{Causal-QA}}$ are trained on the subset of clips with a Causal-QA bank ($\mathcal{D}_{\text{train}}^{\text{QA-only}}$), while the remaining arms use the full chain set, and evaluation is identical across all arms. 
%
%Because the Causal-QA reward requires a QA bank, all four arms are trained on the same subset of clips for which a bank exists ($\mathcal{D}_{\text{train}}^{\text{QA-only}}$), so that the only variable across arms is the reward composition. Evaluation is identical across arms and uses the full test sets.
%
Results are reported in Table~\ref{tab:reward_ablation}, with the non-fine-tuned base model as a reference point.

\begin{table}[t]
    \caption{\textbf{Reward ablation.} Performance of GRPO-fine-tuned Qwen3.5-9B (from the base model, no SFT initialization) as we vary the composition of the composite reward (Equation~\ref{eq:reward}). We report the base (non-fine-tuned) Qwen3.5-9B row for reference.}
    \label{tab:reward_ablation}
    \centering
    \small
    \setlength{\tabcolsep}{4pt}
    \begin{tabular}{@{}ccc rr rr@{}}
    \toprule
    \multicolumn{3}{c}{\textbf{Reward}} & \multicolumn{2}{c}{\textbf{CMD-CC}} & \multicolumn{2}{c}{\textbf{SF20K-CC}} \\
    \cmidrule(lr){1-3} \cmidrule(lr){4-5} \cmidrule(lr){6-7}
    F & S & C & SODA-E & CQA & SODA-E & CQA \\
    \midrule
    \multicolumn{7}{l}{\textit{Base (no fine-tuning)}} \\
    -- & -- & -- & 54.6 & 21.5 & 52.7 & 16.5 \\
    \midrule
    \multicolumn{7}{l}{\textit{GRPO fine-tuning}} \\
    $\checkmark$ & $\times$     & $\times$     & 55.4 & 17.7 & 53.9 & 12.9 \\
    $\checkmark$ & $\checkmark$ & $\times$     & \textbf{60.1} & 26.8 & \textbf{58.8} & 17.7 \\
    $\checkmark$ & $\times$     & $\checkmark$ & 57.0 & \textbf{37.8} & 54.9 & \textbf{26.0} \\
    $\checkmark$ & $\checkmark$ & $\checkmark$ & \underline{58.9} & \underline{32.2} & \underline{57.2} & \underline{24.9} \\
    \bottomrule
    \end{tabular}
\end{table}

\noindent\textbf{Interpretation.}
The reward you optimize is the metric you win. Format-only ($\checkmark$F) regresses below the base model on every metric, since it enforces valid JSON but supplies no content signal. Adding $R_{\text{SODA-E}}$ ($\checkmark$F\,$\checkmark$S) yields the largest SODA-Embedding gain (the highest in the table on both datasets: 60.1 on CMD-CC and 58.8 on SF20K-CC), as expected since it is essentially its own loss. Adding $R_{\text{Causal-QA}}$ ($\checkmark$F\,$\checkmark$C) instead maximizes Causal-QA, reaching 37.8 on CMD-CC (the highest Causal-QA of any open-weights configuration we trained), but at lower SODA-Embedding. The full reward ($\checkmark$F\,$\checkmark$S\,$\checkmark$C) is the best-balanced arm, recovering most of the SODA-Embedding of the SODA-only run while retaining most of the Causal-QA of the QA-only run. The two semantic terms are thus complementary rather than redundant, each binding the metric it targets.

\section{Discussion}
\label{sec:discussion}

We introduced \textbf{Causal Chain Generation}, a new task targeting narrative causality in videos, together with two complementary metrics (\textbf{SODA-Embedding}, an embedding-based extension of the SODA framework capturing semantic matching, and \textbf{Causal-QA}, a question-answering probe of causal coherence) and two datasets: \textbf{SF20K-CC}, a small but contamination-controlled diagnostic over full silent films, and \textbf{CMD-CC}, a larger-scale AD-derived dataset with train and test splits. Our analysis reveals that state-of-the-art VLMs struggle with causal coherence, especially over long event sequences. We also established a strong baseline recipe, combining SFT and GRPO with a composite verifiable reward derived from our metrics, as a reference point for future work on this open challenge.

More broadly, the persistent gap between chain quality and causal accuracy suggests that dense captioning and causal understanding remain distinct capabilities, a distinction our task and metrics are explicitly designed to measure and that we hope will guide future work.

\section*{Limitations}

\noindent\textbf{Benchmark scale and domain.}
SF20K-CC is relatively small (20 silent animated short films, 205 chains), so per-split and per-type estimates have limited statistical power. CMD-CC is larger but consists of short clips with truncated narrative arcs.

\noindent\textbf{Metrics depend on auxiliary models.}
SODA-Embedding relies on a frozen text embedder and Causal-QA on a frozen judge LLM. We show that \emph{rankings} are stable across embedders and across judges of 3B--8B parameters (including cross-family), but \emph{absolute} scores are model-dependent, and a 
weak judge ($\leq$0.5B) fails to discriminate. Results should therefore be read as relative comparisons under the disclosed, fixed embedder and judge, not as absolute capabilities.

%\noindent\textbf{Metric--reward coupling.}
%
%Our GRPO reward optimizes the same SODA-Embedding and Causal-QA models used at evaluation, so part of the fine-tuning gains may reflect alignment to the metrics rather than genuinely better chains. We partially mitigate this with the robustness analysis and a fully independent evaluation (a different judge and embedder).

%\paragraph{Metric--reward coupling.}
%
%Our GRPO reward optimizes the same SODA-Embedding and Causal-QA models used at evaluation, so part of the fine-tuning gains may reflect alignment to the metrics rather than genuinely better chains. We mitigate this with the rank-stability analysis (Appendix~\ref{supmat:metric_validation}), the cross-transfer between the two reward terms (Section~\ref{sec:reward_ablation}), and a held-out judge absent from every reward term (Appendix~\ref{supmat:heldout_judge}). The held-out judge is nonetheless an automatic scorer, and a large-scale human evaluation of fine-tuned outputs is left to future work.

\noindent\textbf{Partial human validation.}
We validate two components with human annotators: the metric--human agreement study, and the answerability of Causal-QA questions given the full chain
%and the causality-type labels on a subsample
. However, we do not report human performance on the CCG task itself.
%
%We report the correlation between human judgment and each metric, verify that Causal-QA questions are reliably answerable by humans given the full chain, and validate the causality-type labels on a subsample. However, we do not report human performance on the CCG task itself.

%\noindent\textbf{Constrained proprietary baselines and compute.}
%
%The proprietary models are run under API constraints (a truncated clip with a limited number of frames). Although we show full-video and truncated-clip results are comparable, this comparison is not fully controlled.

\section*{Acknowledgments}
This work was supported by ANR-22-CE23-0007, a Hi!Paris grant and fellowship, and the ANR/France 2030 program (ANR-23-IACL-0005). Additional support was provided by the Institute of Information \& Technology Planning \& Evaluation (IITP) grant funded by the Korean government (MSIT) (No. RS-2024-00457882, National AI Research Lab Project). We were granted access to the High-Performance Computing (HPC) resources of IDRIS under the allocations 2023-AD011014489 made by GENCI. We would like to express our gratitude to all of the annotators for their help.

%\clearpage

% ---- Bibliography ----
\bibliography{main}

\appendix

\clearpage

\section{Source material}
\label{sec:source_material}

Our benchmark consists of 20 high-quality movies curated from the SF20K dataset. These films are selected based on their rich visual narratives and lack of dialog, ensuring that causal reasoning must be derived only from visual cues. Table~\ref{tab:movie_list} lists the metadata for the included films.

\begin{table*}[h]
    \centering
    \caption{\textbf{SF20K-CC Benchmark metadata.} Detailed list of the 20 movies curated from the SF20K dataset, including YouTube IDs, movie title, and synopsis.}
    \small
    \resizebox{2\columnwidth}{!}{%
    \begin{tabular}{@{}lll@{}}
    \toprule
    \textbf{YouTube ID} & \textbf{Movie Title} & \textbf{Synopsis} \\
    \midrule
    qEaoaRpRRzo & \textit{After The Rain} & A dog turn sheep's wool into clouds. \\
    nfK79Xblpus & \textit{Nullarbor} & A man with road rage confronts an elderly driver. \\
    2p1YK2w8j38 & \textit{Boy In The Woods} & A young boy is lost in the forest. \\
    tzJl6XdByZo & \textit{Wild Love} & The heart has its reasons of which nature knows nothing. \\
    koRvRsflL4Y & \textit{The Sea Is Blue} & The ocean gets its color. \\
    nfK79Xblpus & \textit{Nate And John} & A barber and his client form an unlikely bond that develops over their lifetimes. \\
    eTAqEVMbdRg & \textit{Time Chicken} & The chicken or the egg came first. \\
    aU3YHs30gDo & \textit{The Cocoon} & A man cleans a dirty room. \\
    rU8H5T8zZZM & \textit{Follow Your Heart} & A young woman loses her `heart'. \\
    8SPu5uydAts & \textit{Kayak} & On a peaceful summer day, a dad takes his baby son on a kayak day trip. \\
    CIKEdG4mZxI & \textit{Hors Piste} & A man with road rage confronts an elderly driver. \\
    zVW5Do-WZh8 & \textit{Taking Flight} & A young boy spends a day with grandpa. \\
    CmQneYZvTlM & \textit{Tales From The Multiverse} & God tries to create Earth. \\
    \_kYUf-Qncss & \textit{Corporate Monster} & A man sees monsters. \\
    fSW3a0zb0rQ & \textit{Lily Meets Charlie} & An escort meets a client. \\
    VXI0Gn0Pwck & \textit{Swim} & A swimmer tries out. \\
    j01Hg4QJ6NE & \textit{Best Friend} & In a near future, a lonely man is addicted to a product called ``Best Friend'' \\ & & which offers him perfect virtual friends. \\
    XGI-Vac5kQY & \textit{Hot Dog} & LOVE is a four legged word. \\
    CyGGpsbN55A & \textit{Memo} & Louis, a retired man, resists the overprotection of his daughter Nina to keep \\ & & his independence. \\
    XrqSF2OOz\_M & \textit{Paperman} & Paper airplanes reunite two strangers. \\
    \bottomrule
    \end{tabular}
    }
    \label{tab:movie_list}
\end{table*}

\section{Metric implementation details}
\label{sec:metric}

\begin{figure*}[t]
\centering
\includegraphics[width=1.0\linewidth]{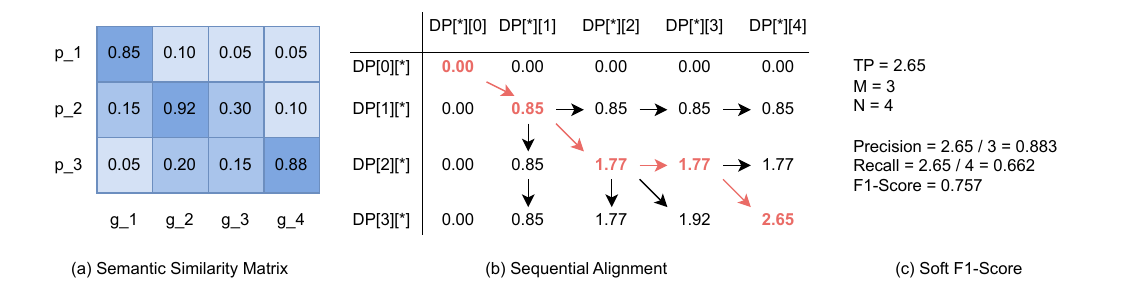}
\caption{\textbf{Illustration of the SODA-Embedding metric.} (a) A semantic similarity matrix is computed (predicted $M=3$, ground-truth $N=4$). (b) An optimal, order-preserving alignment path is found via dynamic programming, yielding a ``soft'' True Positive ($TP$) sum of $2.65$. (c) The final F1-Score is calculated from this $TP$ sum, robustly handling different chain lengths.}
\label{fig:metric_illustration}
\end{figure*}

In this section, we provide the detailed pseudo-code for our proposed SODA-Embedding metric. This metric evaluates the semantic and structural quality of generated causal chains by finding an optimal, order-preserving alignment between the predicted sequence and the ground truth in a semantic embedding space. Figure~\ref{fig:metric_illustration} illustrates the computation on a small example.

\begin{algorithm}[t]
\caption{SODA-Embedding metric calculation}
\label{alg:soda_embedding}
\begin{algorithmic}[1]
\Require Predicted chain $P = (p_1, p_2, \dots, p_m)$
\Require Ground-truth chain $G = (g_1, g_2, \dots, g_n)$
\Require Pre-trained text encoder $\Phi(\cdot)$ (i.e., Qwen3-Embedding-4B)

\State \textbf{Step 1: Semantic Similarity Matrix}
\State $S \leftarrow \text{Matrix}[m][n]$
\For{$i \gets 1$ to $m$}
    \For{$j \gets 1$ to $n$}
        \State $u \gets \Phi(p_i)$
        \State $v \gets \Phi(g_j)$
        \State $S[i][j] \gets \text{CosineSimilarity}(u, v)$
    \EndFor
\EndFor

\State \textbf{Step 2: Sequential Alignment}
\State $DP \leftarrow \text{Matrix}[m+1][n+1]$ initialized to $0$
\For{$i \gets 1$ to $m$}
    \For{$j \gets 1$ to $n$}
        \State $match \gets DP[i-1][j-1] + S[i][j]$
        \State $skip\_p \gets DP[i-1][j]$
        \State $skip\_g \gets DP[i][j-1]$
        \State $DP[i][j] \gets \max(match, skip\_p, skip\_g)$
    \EndFor
\EndFor

\State \textbf{Step 3: Soft F1-Score}
\State $TP_{soft} \gets DP[m][n]$
\State $Precision \gets TP_{soft}\ /\ m$
\State $Recall \gets TP_{soft}\ /\ n$

\If{$Precision + Recall > 0$}
    \State $F1 \gets 2 \cdot Precision \cdot Recall / (Precision + Recall)$
\Else
    \State $F1 \gets 0$
\EndIf

\State \Return $F1$
\end{algorithmic}
\end{algorithm}

\section{Causal-QA examples}
\label{sec:causalqa_examples}

\begin{figure*}[t]
\centering
\resizebox{\textwidth}{!}{%
\begin{tikzpicture}[
  font=\footnotesize, >={Latex[length=1.5mm]},
  frm/.style={inner sep=0pt, draw=black!50, line width=0.4pt},
  ev/.style={rounded corners=2pt, draw=blue!55, fill=blue!6, line width=0.5pt,
             align=center, inner sep=3pt, minimum height=1.25cm, text width=3.0cm},
  ar/.style={-{Latex[length=1.5mm]}, draw=black!60, line width=0.6pt},
  qabox/.style={rounded corners=3pt, draw=darkgreen!55, fill=darkgreen!4, line width=0.6pt,
                align=left, inner sep=5pt, text width=16.4cm},
  tlab/.style={font=\scriptsize\itshape, text=black!70},
  plab/.style={font=\large\bfseries, text=black!70},
]
\def\fh{1.3cm}

% ===== (a) Consequence: marmot, 4 frames, 3-event chain =====
\node[frm] (a1) at (0,0) {\includegraphics[height=\fh]{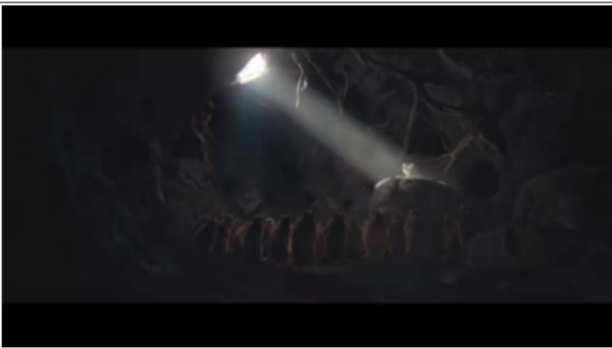}};
\node[frm, right=1mm of a1] (a2) {\includegraphics[height=\fh]{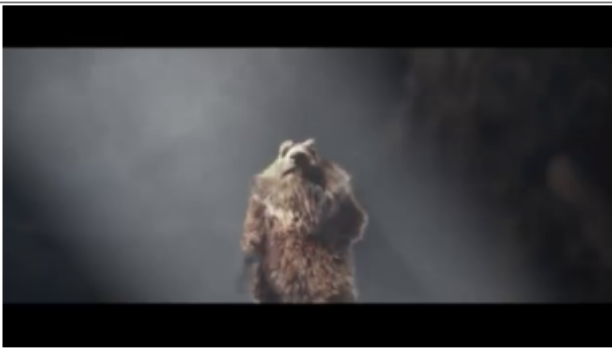}};
\node[frm, right=1mm of a2] (a3) {\includegraphics[height=\fh]{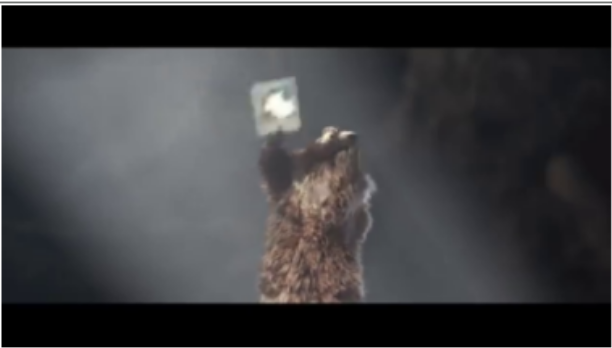}};
\node[frm, right=1mm of a3] (a4) {\includegraphics[height=\fh]{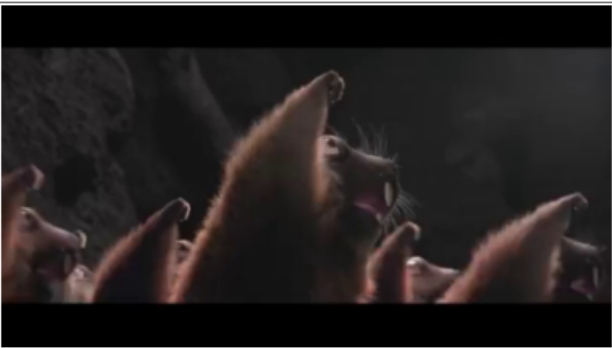}};
\node[plab] at ([xshift=-7mm]a1.west) {(a)};
\node[tlab, right=4mm of a4] {type: Consequence};
\node[ev, anchor=north west] (ea1) at ([yshift=-4mm]a1.south west) {The marmot leader appears on a rock};
\node[ev, right=5mm of ea1] (ea2) {The marmot leader shows a picture of the couple};
\node[ev, right=5mm of ea2] (ea3) {The marmot crowd responds with a scream};
\draw[ar] (ea1) -- (ea2);
\draw[ar] (ea2) -- (ea3);
\node[tlab, anchor=west] at ([xshift=4mm]ea3.east) {ground-truth chain};
\node[qabox, anchor=north west] (qaa) at ([yshift=-4mm]ea1.south west) {%
  \textbf{Q: What is the direct result of the marmot leader showing a picture of the couple?}\\[2pt]
  \begin{tabular}{@{}p{7.7cm}@{\hspace{4mm}}p{7.7cm}@{}}
  A.~The marmot leader disappears. & B.~\textcolor{darkgreen}{\textbf{The marmot crowd screams in response.}}~\textcolor{darkgreen}{\checkmark}\\
  C.~The marmot crowd becomes silent. & D.~The marmot crowd starts digging holes.\\
  E.~This information is not specified. & \\
  \end{tabular}};

% ===== (b) Cause: baby, 3 frames, 4-event chain =====
\node[frm, anchor=north west] (b1) at ([yshift=-8mm]qaa.south west) {\includegraphics[height=\fh]{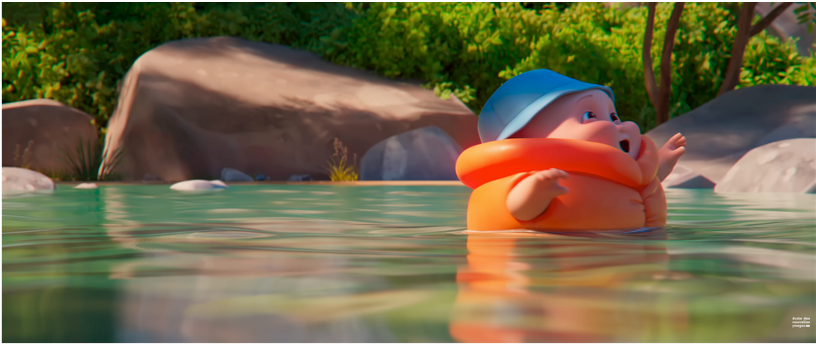}};
\node[frm, right=1mm of b1] (b2) {\includegraphics[height=\fh]{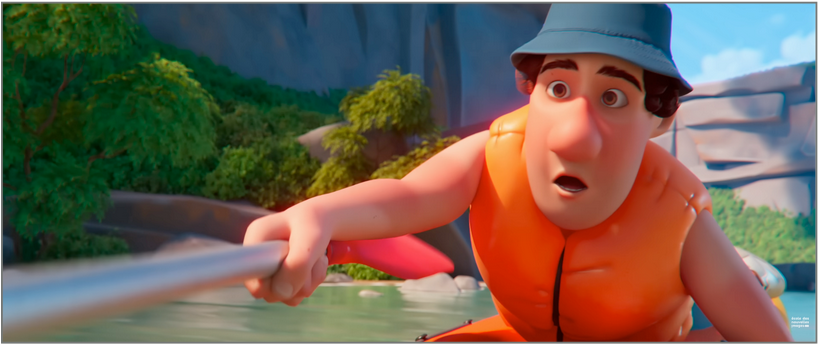}};
\node[frm, right=1mm of b2] (b3) {\includegraphics[height=\fh]{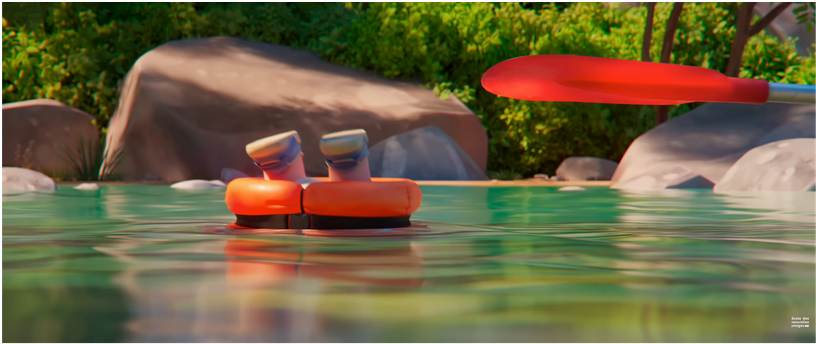}};
\node[plab] at ([xshift=-7mm]b1.west) {(b)};
\node[tlab, right=4mm of b3] {type: Cause};
\node[ev, anchor=north west] (eb1) at ([yshift=-4mm]b1.south west) {The baby floats on the water};
\node[ev, right=5mm of eb1] (eb2) {The father tries to grab the baby with his paddle};
\node[ev, right=5mm of eb2] (eb3) {The father accidentally turns the baby over};
\node[ev, right=5mm of eb3] (eb4) {The baby floats upside down};
\draw[ar] (eb1) -- (eb2);
\draw[ar] (eb2) -- (eb3);
\draw[ar] (eb3) -- (eb4);
\node[qabox, anchor=north west] (qab) at ([yshift=-4mm]eb1.south west) {%
  \textbf{Q: Why did the baby float upside down?}\\[2pt]
  \begin{tabular}{@{}p{7.7cm}@{\hspace{4mm}}p{7.7cm}@{}}
  A.~The baby was initially floating on its back. & B.~The father intentionally flipped the baby.\\
  C.~\textcolor{darkgreen}{\textbf{The baby turned over due to the father's accidental movement.}}~\textcolor{darkgreen}{\checkmark} & D.~The baby rolled over on its own.\\
  E.~This information is not specified. & \\
  \end{tabular}};
\end{tikzpicture}}
\caption{\textbf{Causal-QA examples.} Two items spanning different causality types. For each, the source frames are shown above the ground-truth chain (event boxes read left to right), and the Causal-QA question is answerable only from the intermediate events. The \textcolor{darkgreen}{correct} option is highlighted.}
\label{fig:causalqa}
\end{figure*}

Figure~\ref{fig:causalqa} shows two Causal-QA items spanning different causality types. Each MCQ is answerable only from the intermediate events of the ground-truth chain, so a generated chain that omits or distorts the key causal link leaves the fixed reader LLM unable to recover the correct option.

\section{Training objectives}
\label{supmat:training_losses}

In this section, we detail the two training objectives explored in our work: Supervised Fine-Tuning (SFT) and Group Relative Policy Optimization (GRPO). We provide the mathematical formulations to highlight the fundamental difference in how they optimize the model's reasoning capabilities.

\noindent\textbf{Supervised Fine-Tuning (SFT).}
The SFT baseline optimizes the model to maximize the likelihood of the exact ground-truth token sequence. Given a video $V$ and a target causal chain $C_{gt} = (y_1, y_2, \dots, y_T)$, the loss function is the standard cross-entropy loss:
\begin{equation}
    \mathcal{J}_{SFT}(\theta) = - \frac{1}{T} \sum_{t=1}^{T} \log \pi_\theta(y_t \mid y_{<t}, V)
\end{equation}
SFT penalizes any deviation from the exact phrasing of the ground truth, even if the generated reasoning is semantically correct. This rigidity is suboptimal for open-ended causal reasoning tasks.

\noindent\textbf{Group Relative Policy Optimization (GRPO).}
To address the limitations of SFT, we employ GRPO, which treats the generation as a Reinforcement Learning problem. Instead of a single ground truth, GRPO optimizes the policy $\pi_\theta$ relative to a group of sampled outputs, using a composite reward signal derived from our two metrics (SODA-Embedding and Causal-QA) plus a format reward. For each query $q$, we sample a group of $G$ outputs $\{o_1, o_2, \dots, o_G\}$ from the old policy $\pi_{\theta_{old}}$. The objective function is defined as:
\begin{equation}
\begin{aligned}
    \mathcal{J}_{GRPO}(\theta) = \mathbb{E}_{q \sim \mathcal{D}, \{o_i\} \sim \pi_{\theta_{old}}} \\
    \Big[ \frac{1}{G} \sum_{i=1}^{G} \min \big( r_i(\theta) A_i, \\
    \text{clip}(r_i(\theta), 1-\epsilon, 1+\epsilon) A_i \big) \Big] \\
    - \beta \mathbb{D}_{KL}(\pi_\theta \| \pi_{ref})
\end{aligned}
\end{equation}
where $r_i(\theta) = \frac{\pi_\theta(o_i \mid q)}{\pi_{\theta_{old}}(o_i \mid q)}$ is the probability ratio, $\epsilon$ is the clipping parameter that prevents large policy updates, and $\beta$ controls the strength of the KL-divergence penalty that keeps the model close to the reference model $\pi_{ref}$.

\noindent\textbf{Reward formulation.}
The advantage $A_i$ is computed by normalizing the total reward $R(o_i, C_{gt}, \mathcal{Q}_{gt})$ within the group. To ensure structural validity, semantic matching, and causal coherence, our total reward is composed of three distinct terms:

\begin{equation}
    R(o_i) = R_{\text{format}}(o_i) + R_{\text{soda}}(o_i, C_{gt}) + R_{\text{qa}}(o_i, \mathcal{Q}_{gt})
\end{equation}

\textbf{(i) Format reward ($R_{format}$):}
Since causal reasoning requires structured output to be machine-readable, we impose a strict binary reward for adherence to the JSON schema, defined in Figure~\ref{fig:json_template}.

\begin{figure}[h]
\centering
\begin{lstlisting}[language=json, basicstyle=\ttfamily\small,
                   numbers=left, numberstyle=\tiny, numbersep=5pt,
                   xleftmargin=2em, framexleftmargin=1.5em]
%\begin{lstlisting}[language=json, basicstyle=\ttfamily\small]
{
  "causal_chain": [
    {
      "timestamp": "mm:ss-mm:ss",
      "event": "Event 1 (Start event)"
    },
    {
      "timestamp": "mm:ss-mm:ss",
      "event": "Event 2 (Intermediate)"
    },
    ...
    {
      "timestamp": "mm:ss-mm:ss",
      "event": "Event {num_events} (End event)"
    }
  ]
}
\end{lstlisting}
\vspace{-2mm}
\caption{\textbf{JSON output schema.} The strict JSON formatting template enforced during generation to ensure valid causal chain structure.}
\label{fig:json_template}
\end{figure}

The function verifies that the output $o_i$ is valid JSON and contains a chain with at least 3 events (start, end, and at least one intermediate step).

\begin{equation}
\resizebox{\columnwidth}{!}{$
    R_{format}(o_i) =
    \begin{cases}
    1.0 & \text{if valid JSON} \land \text{len}(o_i) \ge 3 \\
    0.0 & \text{otherwise}
    \end{cases}
$}
\end{equation}

\textbf{(ii) Semantic reward ($R_{soda}$):}
If the format is valid, we evaluate the reasoning quality using the SODA-Embedding metric defined in Section~\ref{subsubsec:sodae}. We scale the metric (originally $[0, 100]$) to the $[0, 1]$ range.
\begin{equation}
\resizebox{\columnwidth}{!}{$
    R_{\text{soda}}(o_i, C_{gt}) =
    \begin{cases}
        \frac{1}{100}\text{SODA}_E(o_i, C_{gt})\ & \text{if valid} \\
        0.0 & \text{otherwise}
    \end{cases}
$}
\end{equation}

\textbf{(iii) Causal-QA reward ($R_{qa}$):}
We pass the generated chain $o_i$ as sole context to the fixed judge LLM (Qwen2.5-3B-Instruct) and measure its accuracy $\text{acc}(o_i, \mathcal{Q})$ on the MCQ bank $\mathcal{Q}$ associated with the clip (Section~\ref{subsubsec:causalqa}). This reward applies to clips that have a Causal-QA bank; for all others it is set to $0$.
\begin{equation}
\resizebox{\columnwidth}{!}{$
    R_{qa}(o_i, \mathcal{Q}) =
    \begin{cases}
    \text{acc}(o_i, \mathcal{Q}) & \text{if valid} \land |\mathcal{Q}| > 0 \\
    0 & \text{otherwise}
    \end{cases}
$}
\end{equation}

\section{Evaluation with a held-out judge}
\label{supmat:heldout_judge}
 
A natural concern with GRPO fine-tuning is that improvements on SODA-Embedding and Causal-QA could partly reflect optimization against the reward signal itself, rather than a genuine improvement in causal-chain reasoning. To assess this, we evaluate all models with a held-out LLM judge (GPT-5.6-Luna) that scores each response on a 0–5 scale and is never used as a reward signal during training.

As shown in Table~\ref{tab:heldout_judge}, the held-out judge score improves monotonically from the pretrained base through SFT to SFT + GRPO on both benchmarks, tracking the trend observed in the reward-coupled metrics. On CMD-CC, the score rises from 2.59 (base) to 2.83 (SFT) to 2.92 (GRPO); on SF20K-CC, it rises from 2.41 to 2.55 to 2.58.

\begin{table}[t]
\centering
\caption{\textbf{Held-out evaluation.} Our two metrics and a held-out judge (GPT-5.6-Luna, 0--5 scale) across the training pipeline. The held-out judge appears in no reward term.}
\label{tab:heldout_judge}
\resizebox{\columnwidth}{!}{%
\begin{tabular}{@{}lrrrrrr@{}}
\toprule
& \multicolumn{3}{c}{\textbf{CMD-CC}} & \multicolumn{3}{c}{\textbf{SF20K-CC}} \\
\cmidrule(lr){2-4} \cmidrule(lr){5-7}
Method & SODA-E & CQA & Judge & SODA-E & CQA & Judge \\
\midrule
Base       & 54.6 & 21.5 & 2.59 & 52.7 & 16.5 & 2.41 \\
SFT        & 62.3 & 24.6 & 2.83 & 56.6 & 19.3 & 2.55 \\
SFT + GRPO & \textbf{63.3} & \textbf{29.4} & \textbf{2.92} & \textbf{57.9} & \textbf{22.1} & \textbf{2.58} \\
\bottomrule
\end{tabular}%
}
\end{table}

\section{Reward ablation across initializations}
\label{supmat:reward_ablation_full}

Table~\ref{tab:reward_ablation} in the main paper reports the reward ablation for GRPO trained directly from the base model. For completeness, Table~\ref{tab:reward_ablation_full} reports the full ablation for both initializations, GRPO from scratch (no SFT) and GRPO on top of the SFT-merged model, under all four reward compositions, using a converged 8{,}192-example training budget. The two initializations specialize in opposite directions. From scratch, the Causal-QA reward drives the highest causal accuracy anywhere (37.8 Causal-QA on CMD-CC, with the SODA-E reward added back recovering chain quality at a small accuracy cost), since the policy chases the judge without a fluency prior. The SFT initialization instead yields the best chain quality (up to 63.3 SODA-E) but caps the Causal-QA gains, as the penalty toward the fluent SFT policy keeps chains well-formed yet prevents aggressive judge-chasing. Across both, format-only GRPO regresses below its initialization on both metrics, confirming that the semantic reward terms, not the format constraint, are what drive the gains.

\begin{table}[t]
    \caption{\textbf{Reward ablation across initializations.} GRPO-fine-tuned Qwen3.5-9B under all four reward compositions ($R_{\text{format}}$ always on; S~$=R_{\text{SODA-E}}$, C~$=R_{\text{Causal-QA}}$), trained from the base model (\emph{scratch}) or on top of the SFT-merged model (\emph{SFT}), at a converged 8{,}192-example budget. The \emph{scratch} block is the from-scratch ablation condensed in Table~\ref{tab:reward_ablation}. Best per column in \textbf{bold}; the from-scratch F+C SF20K cell has no trained eval.}
    \label{tab:reward_ablation_full}
    \centering
    \small
    \setlength{\tabcolsep}{4pt}
    \resizebox{\columnwidth}{!}{%
    \begin{tabular}{@{}ll rr rr@{}}
    \toprule
    & & \multicolumn{2}{c}{\textbf{CMD-CC}} & \multicolumn{2}{c}{\textbf{SF20K-CC}} \\
    \cmidrule(lr){3-4}\cmidrule(lr){5-6}
    Init & Reward & SODA-E & CQA & SODA-E & CQA \\
    \midrule
    \multirow{4}{*}{Scratch}
      & F      & 55.4 & 17.7 & 53.9 & 12.9 \\
      & F+S    & 60.1 & 26.8 & 58.8 & 17.7 \\
      & F+C    & 57.0 & \textbf{37.8} & 54.9 & \textbf{26.0} \\
      & F+S+C  & \underline{58.9} & \underline{32.2} & {57.2} & \underline{24.9} \\
    \midrule
    \multirow{4}{*}{SFT}
      & F      & 57.8 & 18.3 & 53.9 & 14.1 \\
      & F+S    & \textbf{63.3} & 25.5 & \textbf{59.3} & 18.1 \\
      & F+C    & 62.9 & \underline{27.8} & 56.2 & \textbf{24.1} \\
      & F+S+C  & \textbf{63.3} & \textbf{29.4} & \underline{57.9} & \underline{22.1} \\
    \bottomrule
    \end{tabular}%
    }
\end{table}

\section{Test-time scaling}
\label{supmat:tts}

At inference time we can trade compute for quality by sampling several candidate chains from the GRPO-fine-tuned policy and selecting one with a reference-free selector. We use \textit{judge self-consistency}: each candidate is scored by the agreement of the fixed judge LLM's answers on the candidate's own auto-generated MCQ bank, and the highest-scoring candidate is returned. Table~\ref{tab:tts} sweeps the number of samples $k$: both metrics improve monotonically with $k$ and saturate around $k\!=\!16$, lifting CMD-CC SODA-Embedding from 54.9 ($k\!=\!1$) to 60.0 and Causal-QA from 27.5 to 29.9. These runs use the 150-clip CMD-CC subset supported by the selection pipeline, so they are not directly comparable to the main results.

\begin{table}[t]
    \caption{\textbf{Test-time scaling} of GRPO-fine-tuned Qwen3.5-VL-9B: sampling $k$ candidate chains and selecting one by judge self-consistency. Evaluated on the 150-clip CMD-CC subset and the SF20K-CC clips supported by the selection pipeline; not directly comparable to the main tables. Best per column in \textbf{bold}.}
    \label{tab:tts}
    \centering
    \small
    \setlength{\tabcolsep}{6pt}
    \begin{tabular}{@{}rrrrr@{}}
    \toprule
    & \multicolumn{2}{c}{\textbf{CMD-CC}} & \multicolumn{2}{c}{\textbf{SF20K-CC}} \\
    \cmidrule(lr){2-3} \cmidrule(lr){4-5}
    $k$ & SODA-E & CQA & SODA-E & CQA \\
    \midrule
    1  & 54.9 & 27.5 & 53.4 & 18.1 \\
    2  & 56.9 & 27.6 & 53.4 & 18.5 \\
    4  & 58.4 & 28.9 & 54.3 & 20.1 \\
    8  & 59.4 & 29.3 & 53.9 & 18.9 \\
    16 & \textbf{60.0} & \textbf{29.9} & \textbf{54.7} & 18.5 \\
    32 & 59.8 & \textbf{29.9} & \textbf{54.7} & \textbf{20.9} \\
    \bottomrule
    \end{tabular}
\end{table}

\section{Metric validation analysis}
\label{supmat:metric_val}

Table~\ref{tab:metric_synthetic_tests} reports the per-operator success rates summarized in Figure~\ref{fig:metric_validation} (top): for every perturbation operator, text-embedding metrics dominate n-gram metrics, and robustness improves with the scale of the base embedding model.

% \begin{table*}[t]
%     \caption{\textbf{Perturbation analysis} of the SODA-Embedding metric.}
%     \label{tab:metric_synthetic_tests}
%     \centering
%     \small
%     \begin{tabular}{@{}llllrrrr@{}}
%     \toprule
%     \multirow{2}{*}{Metric} & \multirow{2}{*}{Text Similarity}  & \multirow{2}{*}{Base LLM}  & \multirow{2}{*}{Swapping} & \multirow{2}{*}{Shuffling} & Intruder  & Random  \\
%      & &  &  &  & Sentence & Sequence \\
%     \midrule
%     Random & - & - & 50.00 & 50.00 & 50.00 & 50.00 \\
%     \midrule
%     CIDEr & N-gram & - & 45.6 & 56.1 & 67.7 & 75.8 \\
%     BLEU-1 & N-gram &  - & 40.3 & 40.3 & 67.7 & 87.0 \\
%     ROUGE-L & N-gram & - & 35.0 & 49.1 & 70.9 & 90.3 \\
%     METEOR & N-gram & - & 61.4 & 70.1 & 77.4 & 91.9 \\
%     Sentence-BERT & Text embedding & BERT & 61.4 & 61.4 & 83.8 & 96.7 \\
%     Qwen3-Embedding-0.6B & Text embedding & Qwen3 & \textbf{71.9} & 71.9 & 85.4 & \textbf{100.0} \\
%     Qwen3-Embedding-4B & Text embedding & Qwen3 & \textbf{71.9} & \textbf{77.1} & \textbf{87.0} & \textbf{100.0} \\
%     \midrule
%     Oracle & - & - & 100.0 & 100.0 & 100.0 & 100.0 \\
%     \bottomrule
%     \end{tabular}
% \end{table*}

\begin{table}[t]
    \caption{\textbf{Perturbation analysis} of the SODA-Embedding metric.}
    \label{tab:metric_synthetic_tests}
    \centering
    \footnotesize
    % \small
     \resizebox{\linewidth}{!}{  
    \begin{tabular}{@{}llllrrrr@{}}
    \toprule
    \multirow{2}{*}{Metric} & \multirow{2}{*}{Sim}  & \multirow{2}{*}{LLM}  & \multirow{2}{*}{Swapping} & \multirow{2}{*}{Shuffling} & Intruder  & Random  \\
     & &  &  &  & Sentence & Sequence \\
    \midrule
    Random & - & - & 50.00 & 50.00 & 50.00 & 50.00 \\
    \midrule
    CIDEr & N & - & 45.6 & 56.1 & 67.7 & 75.8 \\
    BLEU-1 & N &  - & 40.3 & 40.3 & 67.7 & 87.0 \\
    ROUGE-L & N & - & 35.0 & 49.1 & 70.9 & 90.3 \\
    METEOR & N & - & 61.4 & 70.1 & 77.4 & 91.9 \\
    Sentence-BERT & T & BERT & 61.4 & 61.4 & 83.8 & 96.7 \\
    Qwen3-Embedding-0.6B & T & Qwen3 & \textbf{71.9} & 71.9 & 85.4 & \textbf{100.0} \\
    Qwen3-Embedding-4B & T & Qwen3 & \textbf{71.9} & \textbf{77.1} & \textbf{87.0} & \textbf{100.0} \\
    \midrule
    Oracle & - & - & 100.0 & 100.0 & 100.0 & 100.0 \\
    \bottomrule
    \end{tabular}
    }
\end{table}

%Table~\ref{tab:metric_corr_human} reports the full pairwise meta-evaluation summarized in Figure~\ref{fig:metric_validation} (bottom). Over 256 stratified A/B pairs (128 CMD-CC, 128 SF20K), we report each metric's agreement with the annotator's preference, a 95\% bootstrap confidence interval, and the Kendall $\tau = 2\,\text{acc} - 1$ on decided pairs. Every embedding-based metric (77--79\%) significantly outranks every lexical metric. The decisive comparison is SODA-E against its exact lexical ablation SODA-C (same SODA alignment, METEOR scorer instead of embeddings): 79.2\% versus 66.1\% ($p<0.001$, paired McNemar), confirming that the embedding scorer drives the agreement. Within the embedding cluster, SODA-E is statistically tied with a naive Qwen cosine, MiniLM-based SODA, and BERTScore, so we do not claim superiority over other embedding metrics, only over lexical ones. The annotator is a strong LLM (GPT-5.4) used as a reproducible proxy for a human; its three independent runs agree almost perfectly (Fleiss' $\kappa=0.88$), but it remains a pilot rather than a human study.

Table~\ref{tab:metric_corr_human} reports the full pairwise meta-evaluation summarized in Figure~\ref{fig:metric_validation} (bottom). Over 256 stratified A/B pairs (128 CMD-CC, 128 SF20K), we report each metric's agreement with the majority human preference, a 95\% bootstrap confidence interval, and Kendall's $\tau = 2\,\text{acc} - 1$ on decided pairs. All embedding-based metrics (76--79\%) outrank all lexical metrics. The decisive comparison is SODA-E against its exact lexical ablation SODA-C, which retains the SODA alignment but substitutes METEOR for the embedding scorer: 79.2\% versus 66.1\% ($p<0.001$, paired McNemar), indicating that the embedding scorer accounts for the gain. Across SODA-E variants, agreement increases with the capacity of the underlying encoder: 76.2\% with MiniLM-L6-v2, 78.5\% with Qwen3-Embedding-0.6B, and 79.2\% with Qwen3-Embedding-4B, which we adopt. Each pair was judged independently by three human annotators, with strong inter-annotator agreement (Fleiss' $\kappa=0.881$).

\begin{table}[t]
    \caption{\textbf{Agreement with human judgment.} Pairwise meta-evaluation over 256 A/B pairs: a metric \emph{agrees} when the prediction it scores higher is the one annotators prefer. We report agreement, a 95\% bootstrap confidence interval, and Kendall $\tau = 2\,\text{acc} - 1$ on decided pairs. SODA-C is the lexical ablation of SODA-Embedding (same alignment, METEOR scorer). Embedding metrics are in \textbf{bold}.}
    \label{tab:metric_corr_human}
    \centering
    \small
    \setlength{\tabcolsep}{5pt}
    \begin{tabular}{@{}llcc@{}}
    \toprule
    Metric & Family & Agree. & $\tau$ \\
    \midrule
    \textbf{SODA-E (Qwen-4B)} & \textbf{embed} & \textbf{79.2} & .584 \\
    \textbf{SODA-E (Qwen-0.6B)} & \textbf{embed} & \textbf{78.5} & .570 \\
    \textbf{BERTScore}        & \textbf{embed} & \textbf{77.3} & .547 \\
    \textbf{SODA-E (MiniLM)}  & \textbf{embed} & \textbf{76.2} & .523 \\
    \midrule
    ROUGE-L & lexical & 72.0 & .441 \\
    SODA-C  & lexical & 66.1 & .323 \\
    CIDEr   & lexical & 63.9 & .278 \\
    METEOR  & lexical & 61.8 & .236 \\
    \bottomrule
    \end{tabular}
\end{table}

\noindent\textbf{Robustness to the underlying models.}
Table~\ref{tab:metric_robustness} details the rank-stability experiment summarized in Section~\ref{subsec:metric_validation}. We score a fixed panel of nine systems on both test sets, then vary only the metric's auxiliary model: the SODA-Embedding text embedder (Qwen3-Embedding-0.6B/4B/8B and the cross-family all-MiniLM-L6-v2) or the Causal-QA judge (Qwen2.5-Instruct-0.5B/3B/7B, the cross-family Llama-3.1-8B-Instruct, and the recent cross-family Gemma-3-12B-it). For each variant we report how closely the system ranking it induces matches the ranking from our default model (Qwen3-Embedding-4B for SODA-E, Qwen2.5-3B-Instruct for Causal-QA), as the percentage of system pairs ordered identically, the Spearman $\rho$, and the Kendall $\tau$. Rather than collapsing these into a mean, which would hide the single weak model, we list every variant separately.

The rankings are preserved across both model size and model family. SODA-Embedding agrees with its 4B reference at $\ge 91\%$ (Spearman $\rho\ge 0.91$) for every embedder, including the cross-family all-MiniLM-L6-v2, even though the latter shifts absolute scores down by roughly 13 points. Causal-QA is equally stable whenever the judge is competent: the 7B Qwen judge and the two cross-family judges (Llama-3.1-8B and the 2025 Gemma-3-12B) reproduce the 3B reference ranking at $\ge 85\%$ agreement. The lone exception is the 0.5B judge, whose accuracy is nearly flat across systems (close to the one-in-five chance band), so it cannot discriminate and falls to near-chance agreement ($\rho\approx 0$) on the smaller SF20K set. We therefore adopt Qwen3-Embedding-4B and Qwen2.5-3B-Instruct.

\begin{table}[t]
    \caption{\textbf{Metric robustness.} Per-model stability of system rankings when only the metric's underlying model is varied. For each variant we report agreement with the default-model ranking (\% of system pairs ordered identically), Spearman $\rho$, and Kendall $\tau$, on both test sets. References: Qwen3-Embedding-4B (SODA-E), Qwen2.5-3B-Instruct (Causal-QA). $\dagger$: cross-family model.}
    \label{tab:metric_robustness}
    \centering
    \small
    \setlength{\tabcolsep}{4pt}
    \resizebox{\columnwidth}{!}{%
    \begin{tabular}{@{}llrrrrrr@{}}
    \toprule
    & & \multicolumn{3}{c}{\textbf{CMD-CC}} & \multicolumn{3}{c}{\textbf{SF20K}} \\
    \cmidrule(lr){3-5}\cmidrule(lr){6-8}
    Metric & Model & Agr. & $\rho$ & $\tau$ & Agr. & $\rho$ & $\tau$ \\
    \midrule
    \multirow{3}{*}{SODA-E}
      & 0.6B            & 97  & .97  & .94  & 97  & .98  & .94 \\
      & 8B              & 100 & .99  & 1.00 & 100 & 1.00 & 1.00 \\
      & MiniLM$^\dagger$ & 94  & .95  & .89  & 91  & .91  & .83 \\
    \midrule
    \multirow{4}{*}{Causal-QA}
      & 0.5B               & 89  & .92  & .78  & 52  & $-$.02 & .03 \\
      & 7B                 & 97  & .98  & .94  & 89  & .90  & .77 \\
      & Gemma-12B$^\dagger$ & 94  & .95  & .89  & 85  & .82  & .71 \\
      & Llama-8B$^\dagger$  & 100 & 1.00 & 1.00 & 85  & .84  & .71 \\
    \bottomrule
    \end{tabular}%
    }
\end{table}

\section{Baseline setup}
\label{supmat:baseline_setup}

We evaluate four families of systems with SODA-Embedding and Causal-QA: (i) end-to-end VLMs of varying scale (the Qwen3.5-VL family, 0.8B--27B~\cite{qwen3vl}, GLM-4.6V-flash~\cite{glmv}, and InternVL-3.5-8B~\cite{internvl35}); (ii) proprietary models (GPT-5.4-nano/mini~\cite{gpt5}); (iii) causal-reasoning baselines (a chain-of-thought ``thinking'' variant, MECD~\cite{mecd}, and TEMPURA-3B~\cite{tempura}); and (iv) two-stage caption-then-reason pipelines. Each model is run under one of two input conditions: the full video sampled at 1 FPS, or a truncated clip centered on the causal chain from which we sample 32 frames ($\approx$1 FPS over the clip). Open-weights models use the full video; the proprietary GPT-5.4 and Gemini models are restricted to the truncated-clip condition by API constraints (the two conditions are comparable, see Appendix~\ref{supmat:fullvstrunc}). All models receive the clip's subtitle transcript and the ground-truth number of events. We further report three trivial references: \textit{Random}, a shuffled chain over a generic event vocabulary; \textit{Guessing}, a text LLM given only the query; and \textit{Blind}, a text LLM given only the subtitle transcript and no frames. Metric values use Qwen3-Embedding-4B (SODA-Embedding) and Qwen2.5-3B-Instruct (Causal-QA), in line with Section~\ref{subsec:metric_definition}.

\section{Baseline scaling}
\label{supmat:scaling}

\begin{figure}[!htbp]
    \centering
    \includegraphics[width=\linewidth]{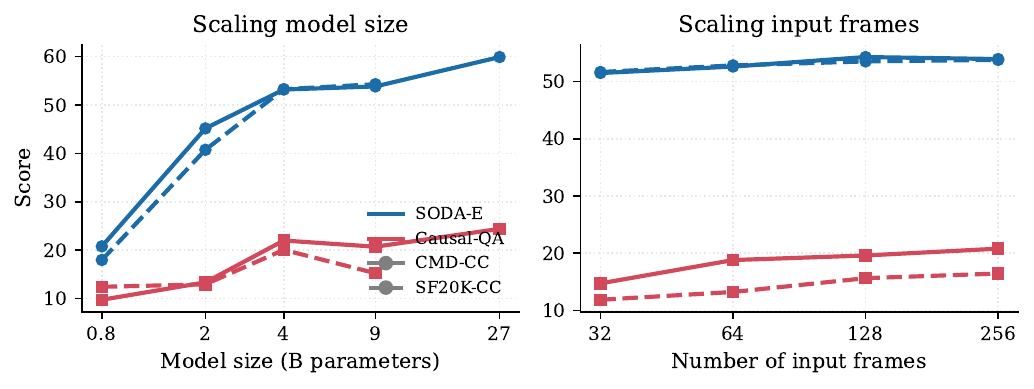}
    \caption{\textbf{Baseline scaling} for the Qwen3.5-VL family on the full video. \textit{Left:} vs.\ model size (0.8B--27B); both metrics rise steeply up to 4B and then saturate. \textit{Right:} vs.\ number of input frames for the 9B model; Causal-QA keeps rising while SODA-Embedding saturates around 128 frames.}
    \label{fig:scaling}
\end{figure}

Figure~\ref{fig:scaling} sweeps two axes for the Qwen3.5-VL family on the full video. With model size, performance rises steeply up to 4B and then largely saturates---SODA-Embedding climbs from 18.0/20.8 (0.8B) to 53.3/53.2 (4B) on SF20K-CC/CMD-CC, with only marginal gains beyond. With the number of input frames, returns are roughly logarithmic: going from 32 to 256 frames adds $+6.1$ Causal-QA points on CMD-CC and $+4.6$ on SF20K-CC, whereas SODA-Embedding saturates around 128 frames. Causal-QA is thus the more frame-hungry axis.

\section{Full vs.\ truncated video}
\label{supmat:fullvstrunc}

Our proprietary baselines are restricted by API limits to a truncated clip centered on the causal chain (32 sampled frames), whereas open-weights models see the full video at 1 FPS. To verify this does not confound the comparison, we run the same backbone (Qwen3.5-VL-9B, with subtitles) under both conditions (Table~\ref{tab:fullvstrunc}). The two conditions differ by at most $\sim$2 points on either metric and either test set, and leave the system ranking unchanged. We therefore treat the full-video and truncated-clip conditions as comparable and place all baselines in a single table (Table~\ref{tab:baselines}).

\begin{table}[t]
    \caption{\textbf{Full vs.\ truncated video} for Qwen3.5-VL-9B (with subtitles). Sampling the full video at 1 FPS and sampling 32 frames from a clip centered on the chain give near-identical scores.}
    \label{tab:fullvstrunc}
    \centering
    \small
    \setlength{\tabcolsep}{6pt}
    \resizebox{\columnwidth}{!}{%
    \begin{tabular}{@{}lrrrr@{}}
    \toprule
    & \multicolumn{2}{c}{\textbf{CMD-CC}} & \multicolumn{2}{c}{\textbf{SF20K-CC}} \\
    \cmidrule(lr){2-3} \cmidrule(lr){4-5}
    Condition & SODA-E & CQA & SODA-E & CQA \\
    \midrule
    Full video (1 FPS)          & 53.8 & 20.7 & 54.3 & 15.3 \\
    Truncated clip (32 frames)  & 54.6 & 21.5 & 52.7 & 16.5 \\
    \midrule
    $\Delta$ & $+0.8$ & $+0.8$ & $-1.6$ & $+1.2$ \\
    \bottomrule
    \end{tabular}%
    }
\end{table}

\section{Difficulty breakdown}
\label{supmat:breakdown}

\begin{figure*}[t]
    \centering
    \includegraphics[width=\linewidth]{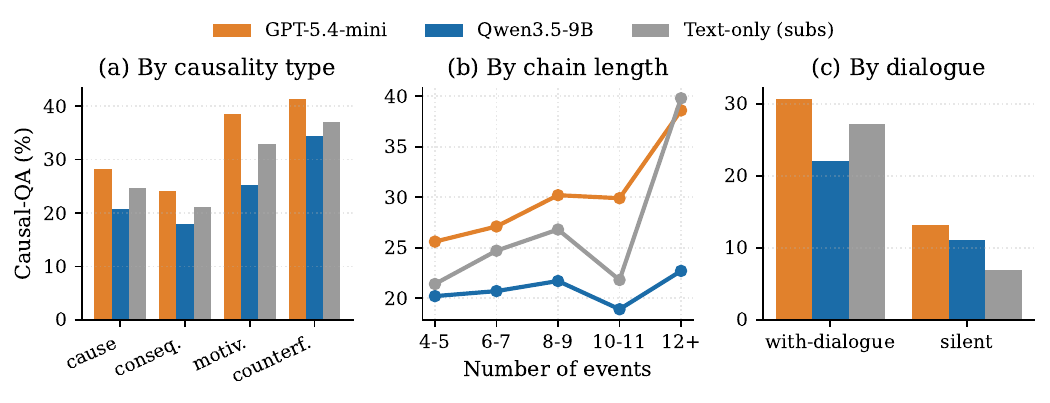}
    \caption{\textbf{Where the difficulty lives.} Causal-QA accuracy on CMD-CC sliced by (a) causality type, (b) gold chain length, and (c) dialogue availability, for GPT-5.4-mini, Qwen3.5-9B (full video with subtitles), and a text-only control that sees the subtitles but no frames. Dialogue is the sharpest axis: silent movies are significantly harder to understand.}
    \label{fig:breakdown}
\end{figure*}

Figure~\ref{fig:breakdown} slices CMD-CC Causal-QA along three axes, comparing GPT-5.4-mini, Qwen3.5-VL-9B with subtitles, and a text-only control that sees the subtitles but no frames. Causality types follow a stable hardness ordering across all models, with \textit{consequence} the hardest and \textit{motivation} the easiest among the well-populated types, while \textit{counterfactual} appears easy but rests on very few questions (29 on CMD-CC) and is exploitable by the judge, so we read it with caution. Performance is essentially flat in chain length, with neither metric decaying from short to long chains. Most strikingly, dialogue availability dominates: every model that sees subtitles scores $+11$ to $+20$ Causal-QA points higher on with-dialogue clips than on silent ones, and on with-dialogue clips the text-only control rivals the video model. Stripping the subtitle track from Qwen3.5-VL-9B erases this advantage entirely ($-4.2$ on CMD-CC), confirming it comes from the subtitle text rather than from dialogue clips being intrinsically easier to watch.

%\section{Performance by narrative causality type}
\section{Causal chain type distribution}
\label{supmat:causality_types}

Table~\ref{tab:causality_dist} reports how the four-type taxonomy (Section~\ref{subsec:benchmark_creation}) is distributed across the test sets. Motivational, Psychological, and Physical together cover over 90\% of chains; Enabling is rare and Other is negligible, so the taxonomy fits the data. The split character differs intuitively: dialogue-rich CMD-CC skews toward intent (Motivational + Psychological $\approx$66\%), whereas the whole-film SF20K skews Physical (45\%).

\begin{table}[t]
    \caption{\textbf{Causality-type distribution} across the CCG test sets (\% of chains; CMD-CC over 1{,}031 chains). Per-type labels are annotated by an LLM classifier.}
    \label{tab:causality_dist}
    \centering
    \small
    \setlength{\tabcolsep}{8pt}
    \begin{tabular}{@{}lrr@{}}
    \toprule
    Causality type & CMD-CC & SF20K-CC \\
    \midrule
    Motivational  & 35.8\% & 33.7\% \\
    Psychological & 30.6\% & 14.7\% \\
    Physical      & 24.2\% & 45.3\% \\
    Enabling      & \phantom{0}8.3\% & \phantom{0}5.3\% \\
    Other         & \phantom{0}1.2\% & \phantom{0}1.1\% \\
    \bottomrule
    \end{tabular}
\end{table}

%Broken down by type, chain \emph{description} quality (SODA-Embedding) is largely type-agnostic for competent models: the per-type spread at a fixed model is only $\sim$1--4 points, and what moves the score is model capacity rather than the type of causal relation (the SF20K size scan climbs steeply within every type). Chain \emph{reasoning} (Causal-QA), in contrast, mildly favors intent-based types: open VLMs score a few points higher on Motivational/Psychological than on Physical on both splits, since Physical chains are easy to describe but their questions probe precise action ordering that models often get wrong.

\section{Success and failure cases}
\label{supmat:qualitative}

\begin{figure*}[htbp]
    \centering
    \tiny
    \newcolumntype{Y}{>{\centering\arraybackslash}X}
    \renewcommand{\arraystretch}{1.4}
    \setlength{\tabcolsep}{3pt}

    \begin{tabularx}{\textwidth}{@{} l YYY @{}}
        & \includegraphics[width=\linewidth]{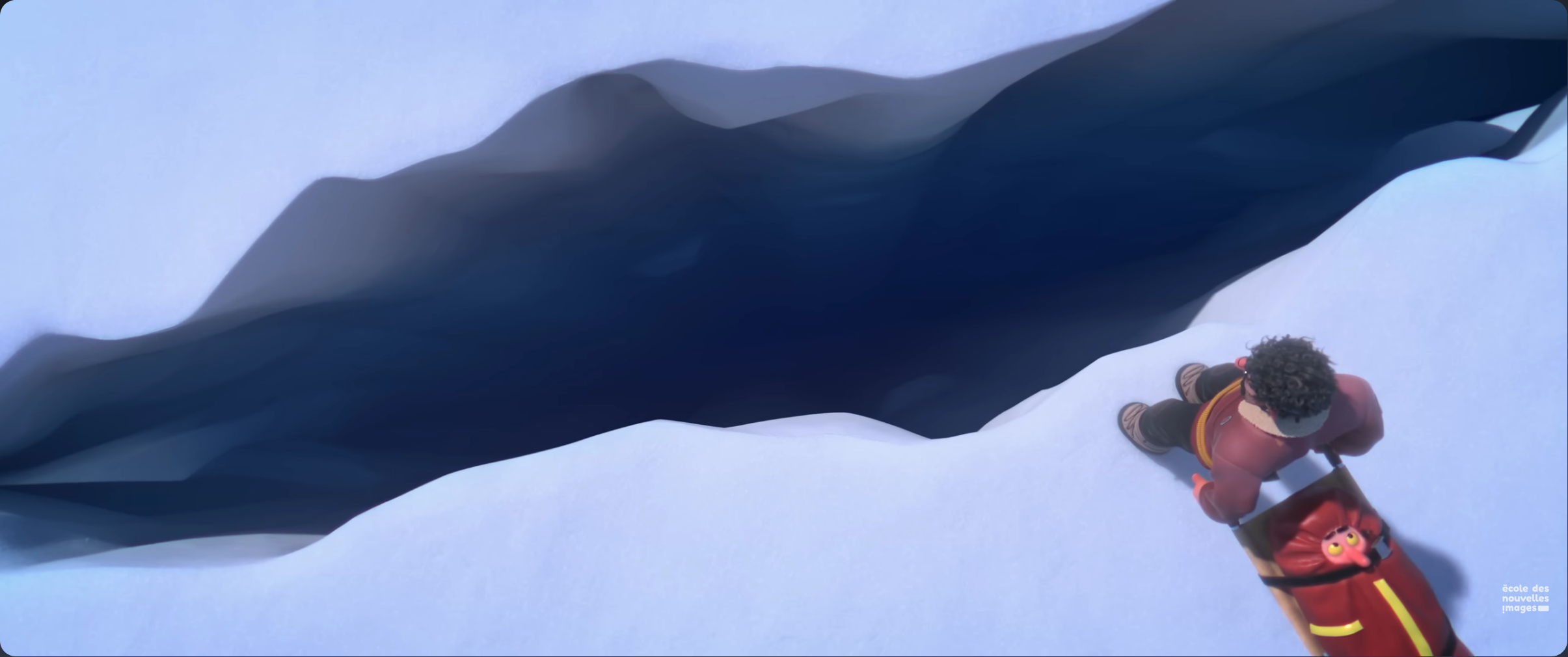} &
          \includegraphics[width=\linewidth]{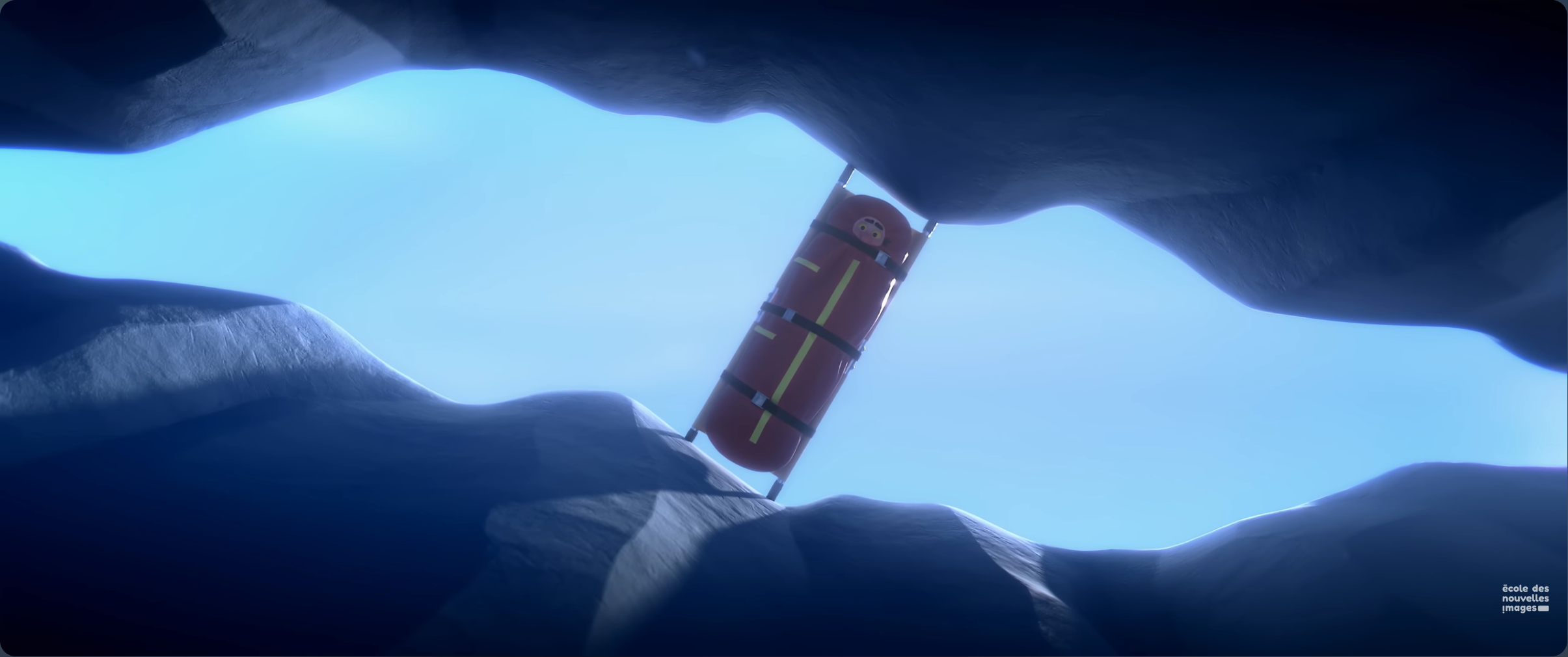} &
          \includegraphics[width=\linewidth]{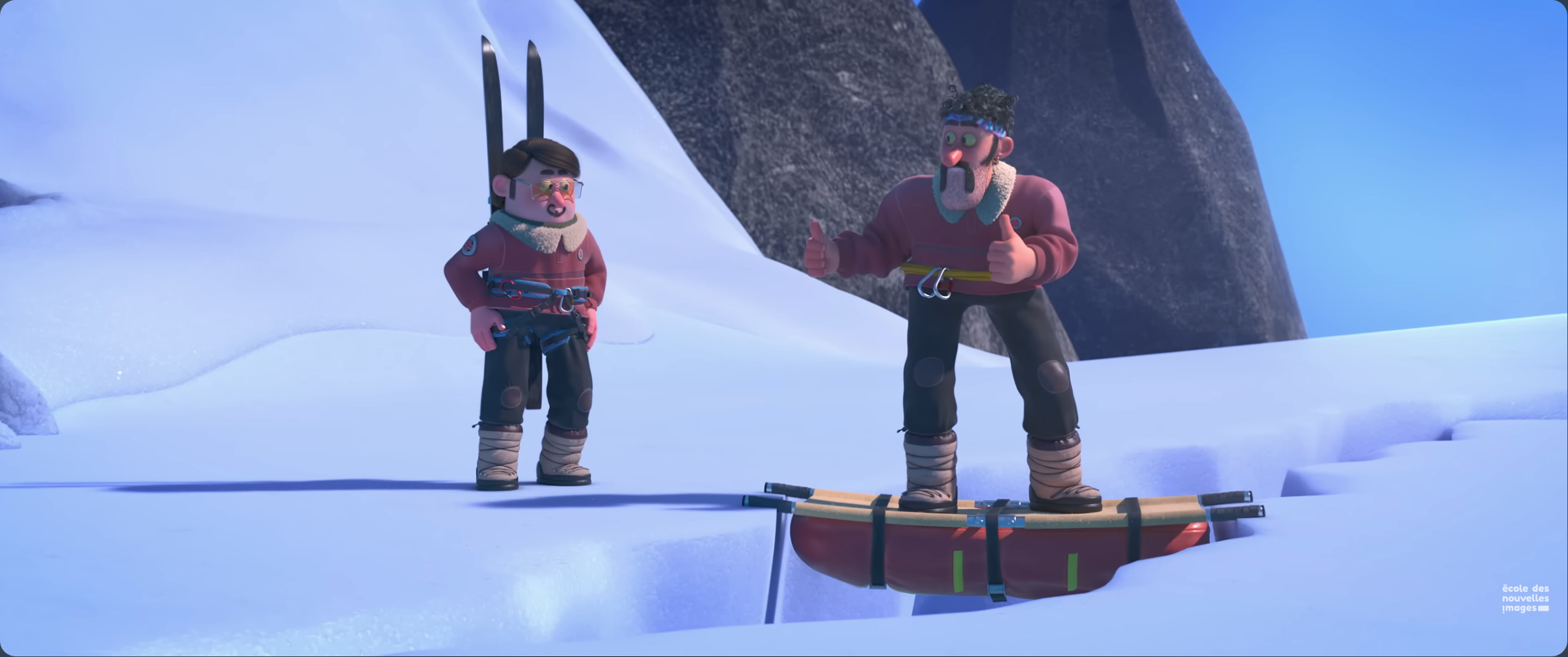} \\
        \midrule
        \textbf{GT} & The rescue workers arrive at a huge gap. &
        \textcolor{darkgreen}{The rescue workers create a bridge with the injured man.} &
        The helicopter falls from the mountain. \\
        \cmidrule(lr){2-4}
        \textbf{Pred} & The rescue workers arrive at a huge gap. &
        \textcolor{darkgreen}{The rescue workers prepare to cross the gap with a stretcher.} &
        The helicopter falls from the mountain. \\
    \end{tabularx}

    \vspace{1.5em}
    \hrule height 1pt
    \vspace{1.5em}

    \begin{tabularx}{\textwidth}{@{} l YYY @{}}
        & \includegraphics[width=\linewidth]{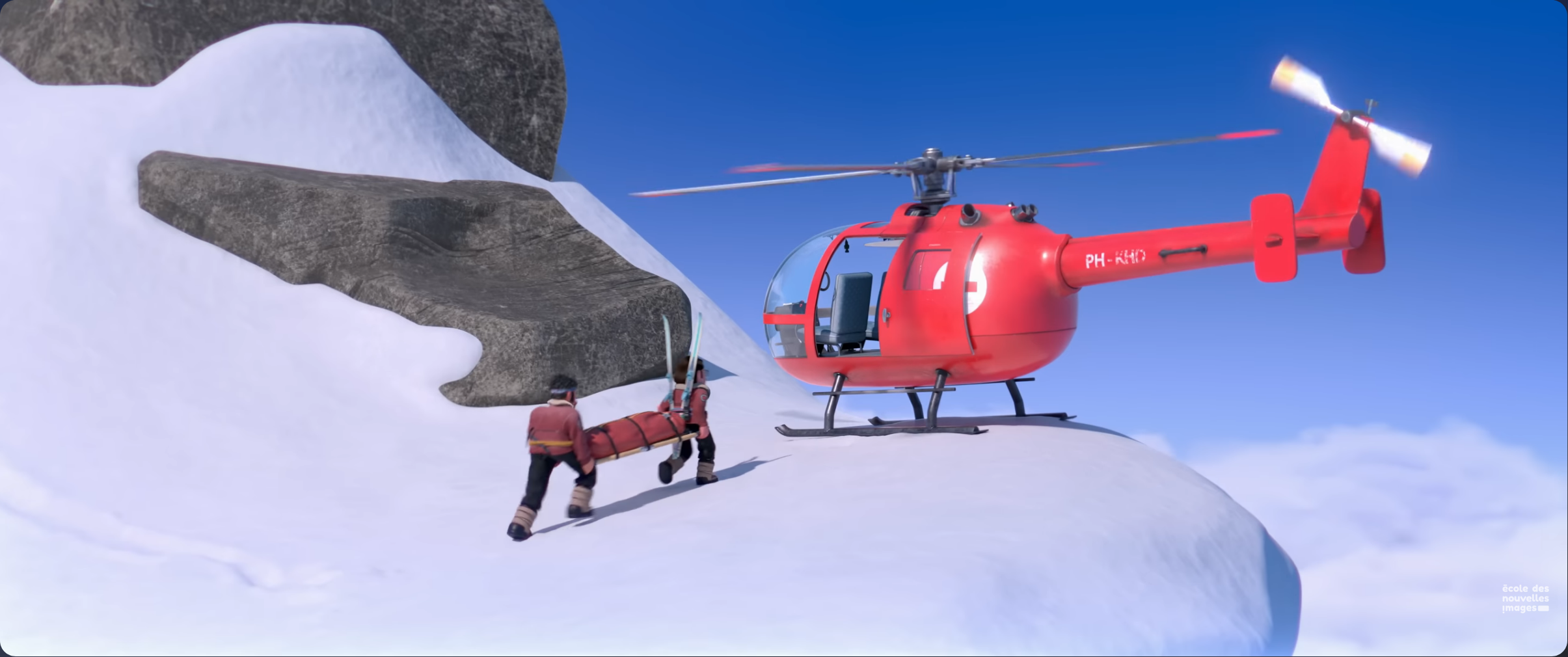} &
          \includegraphics[width=\linewidth]{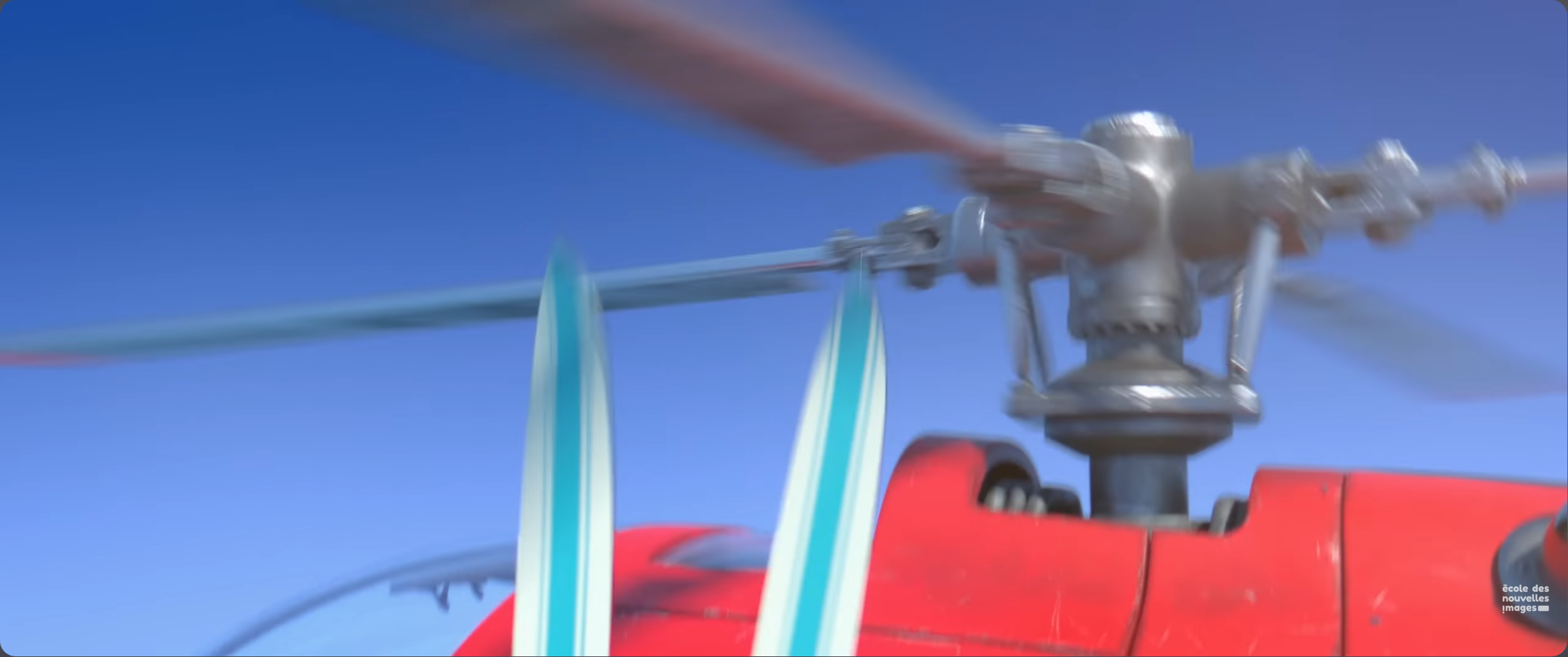} &
          \includegraphics[width=\linewidth]{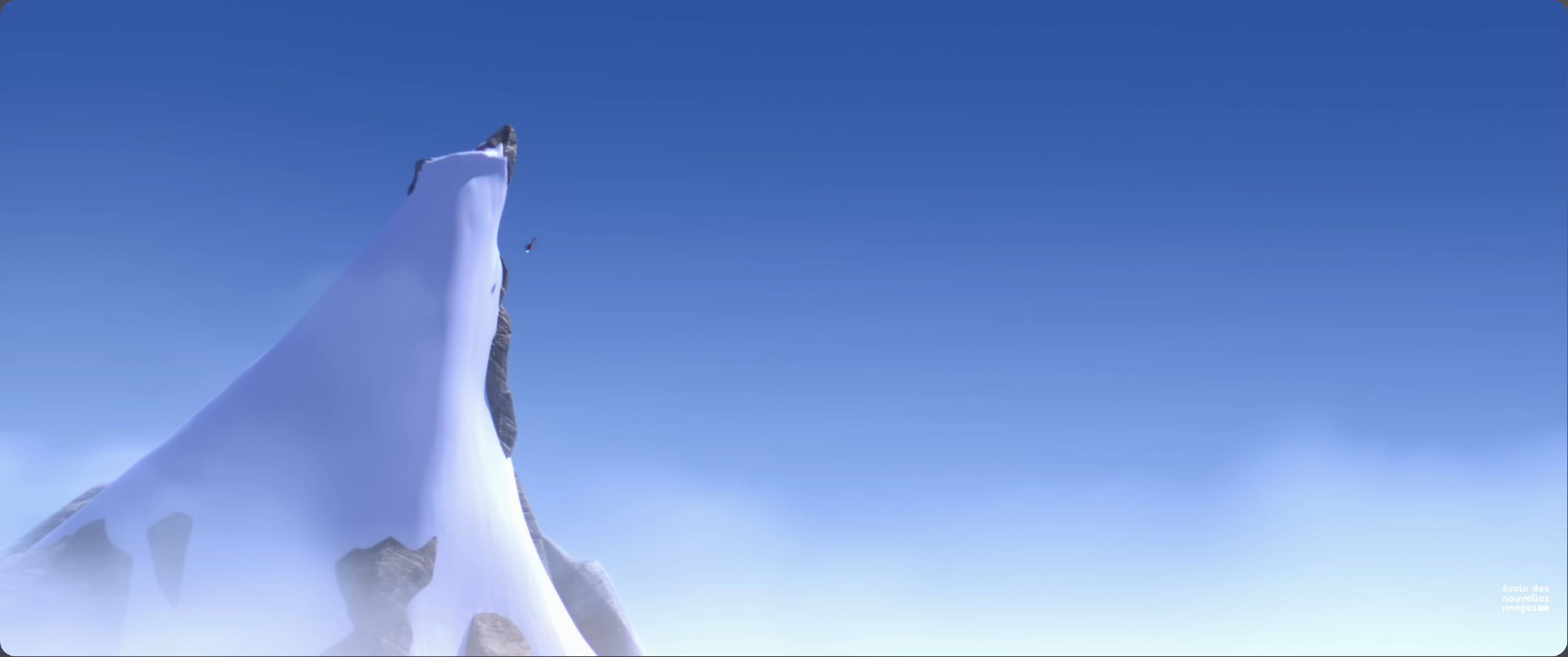} \\
        \midrule
        \textbf{GT} & The rescue worker gets in the helicopter. &
        \textcolor{darkgreen}{The rescue worker hits the helicopter's propellers with his skis.} &
        The helicopter falls from the mountain. \\
        \cmidrule(lr){2-4}
        \textbf{Pred} & The rescue worker gets in the helicopter. &
        \textcolor{firebrick}{The helicopter takes off and approaches the mountain.} &
        The helicopter falls from the mountain. \\
    \end{tabularx}

    \vspace{1.5em}
    \hrule height 1pt
    \vspace{1.5em}

    \begin{tabularx}{\textwidth}{@{} l YYY @{}}
        & \includegraphics[width=\linewidth]{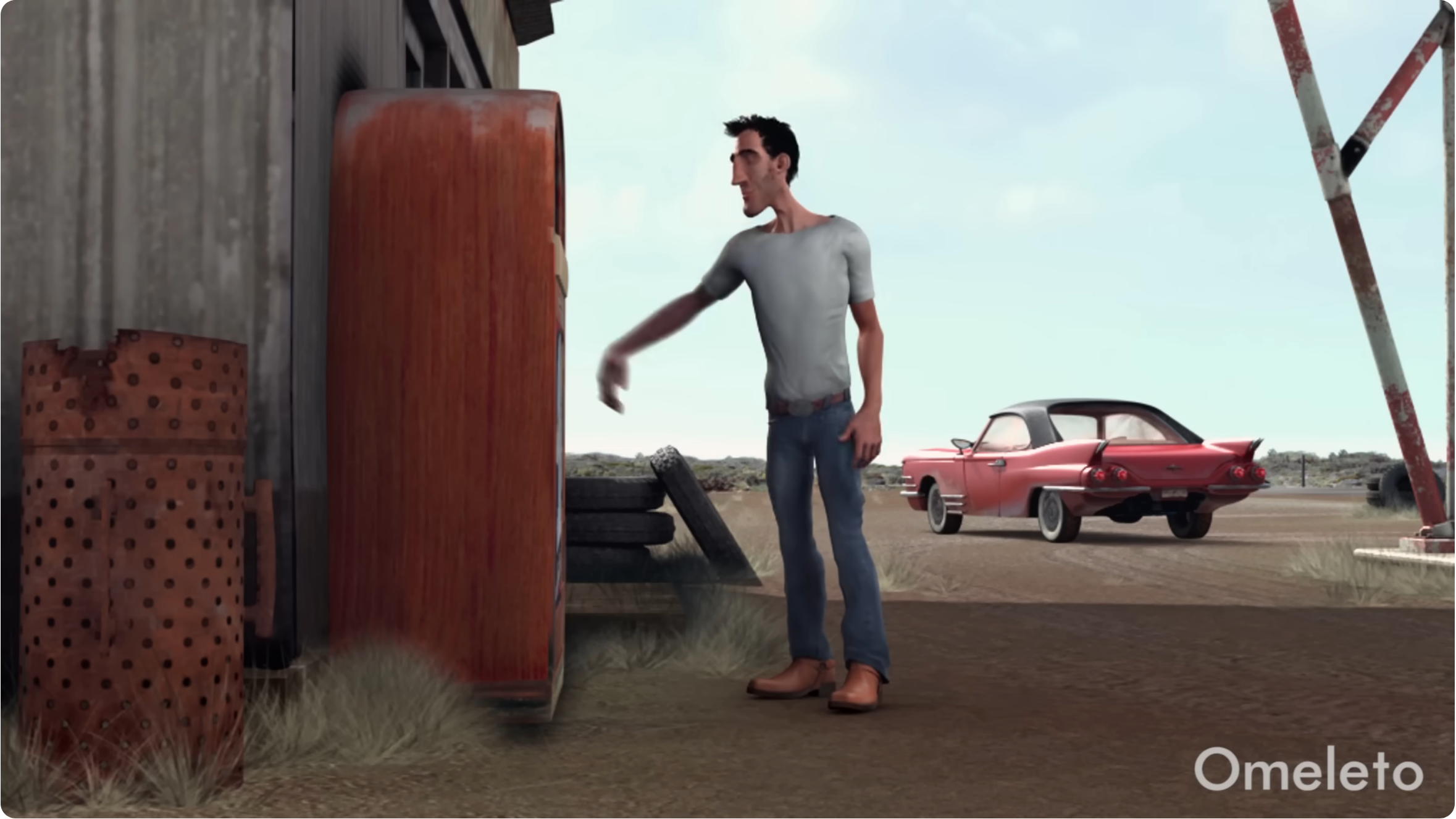} &
          \includegraphics[width=\linewidth]{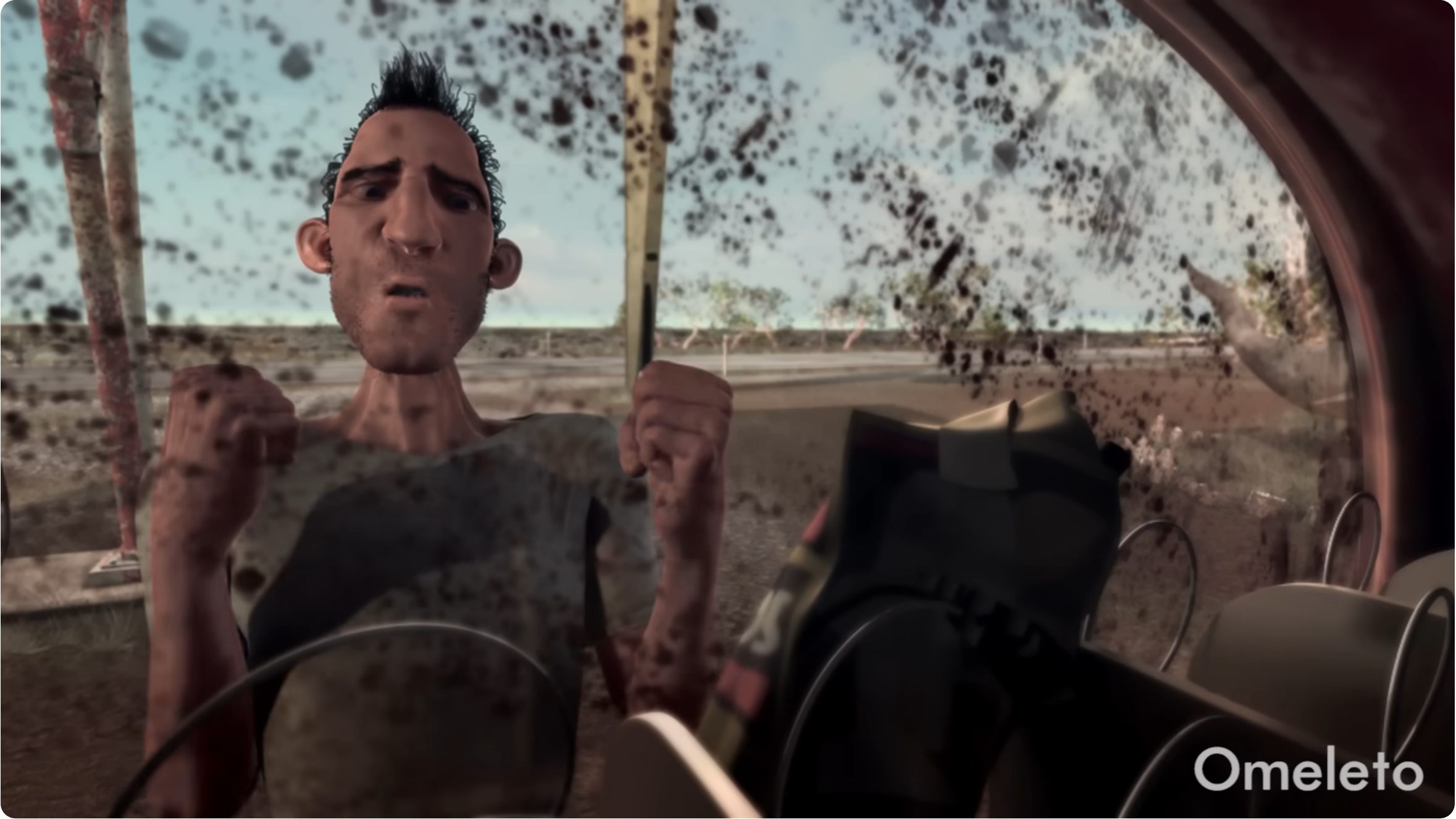} &
          \includegraphics[width=\linewidth]{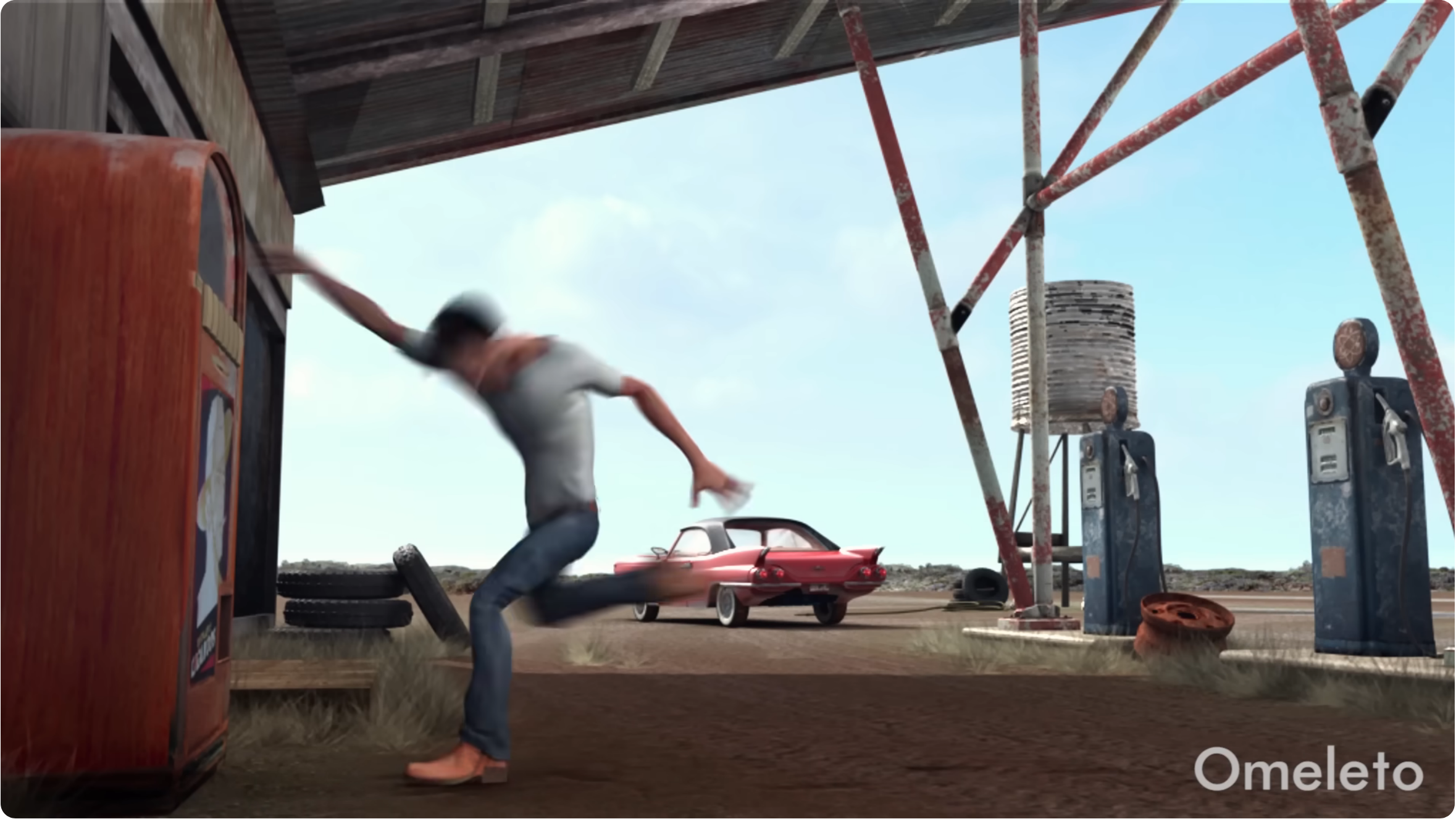} \\
        \midrule
        \textbf{GT} & The man puts a coin in a vending machine. &
        \textcolor{darkgreen}{The cigarettes get stuck in the vending machine.} &
        The man hits the vending machine with a crowbar. \\
        \cmidrule(lr){2-4}
        \textbf{Pred} & The man puts a coin in a vending machine. &
        \textcolor{darkgreen}{The vending machine fails to dispense the item, frustrating the man.} &
        The man hits the vending machine with a crowbar. \\
    \end{tabularx}

    \vspace{1.5em}
    \hrule height 1pt
    \vspace{1.5em}

    \begin{tabularx}{\textwidth}{@{} l YYY @{}}
        & \includegraphics[width=\linewidth]{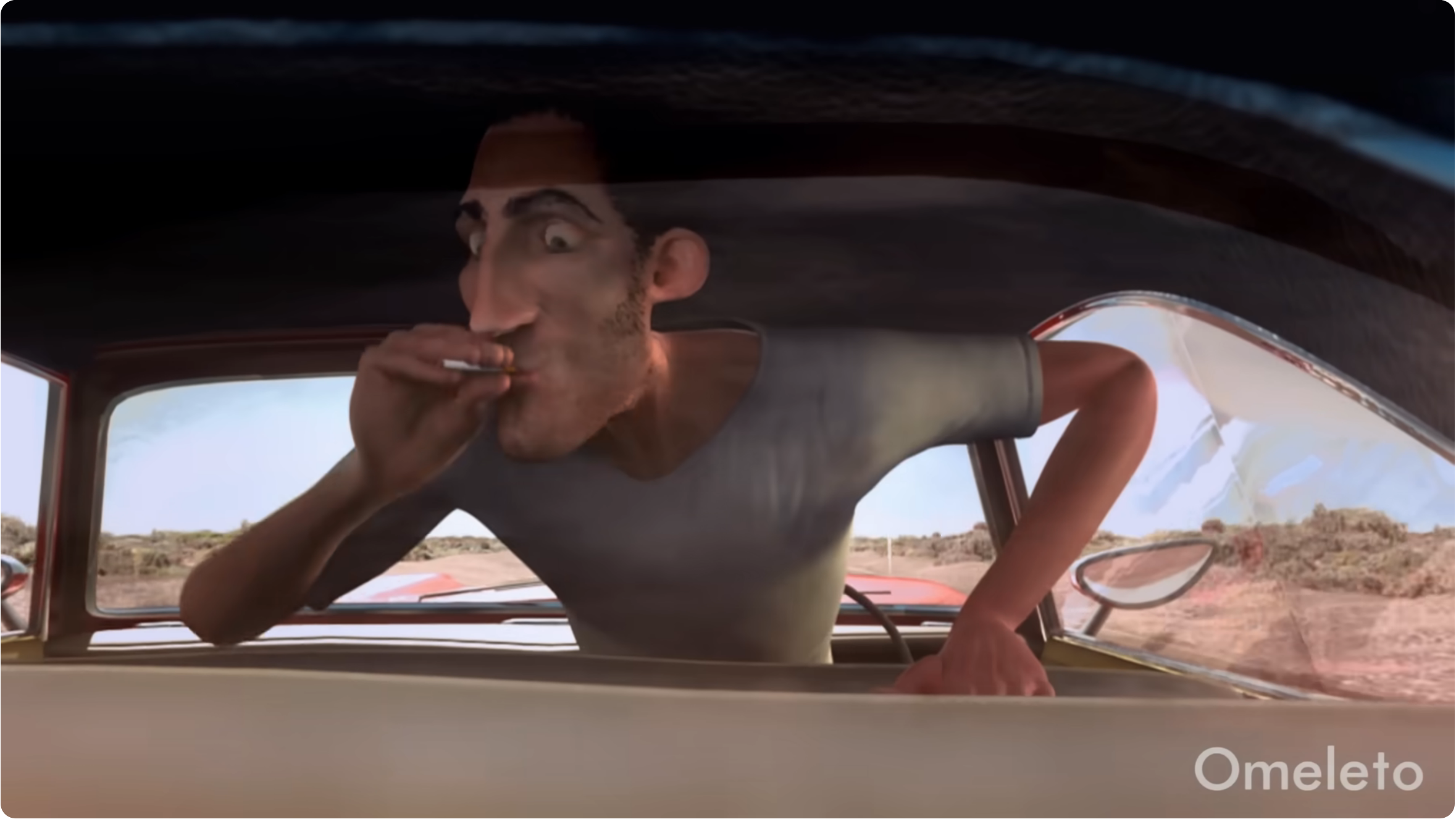} &
          \includegraphics[width=\linewidth]{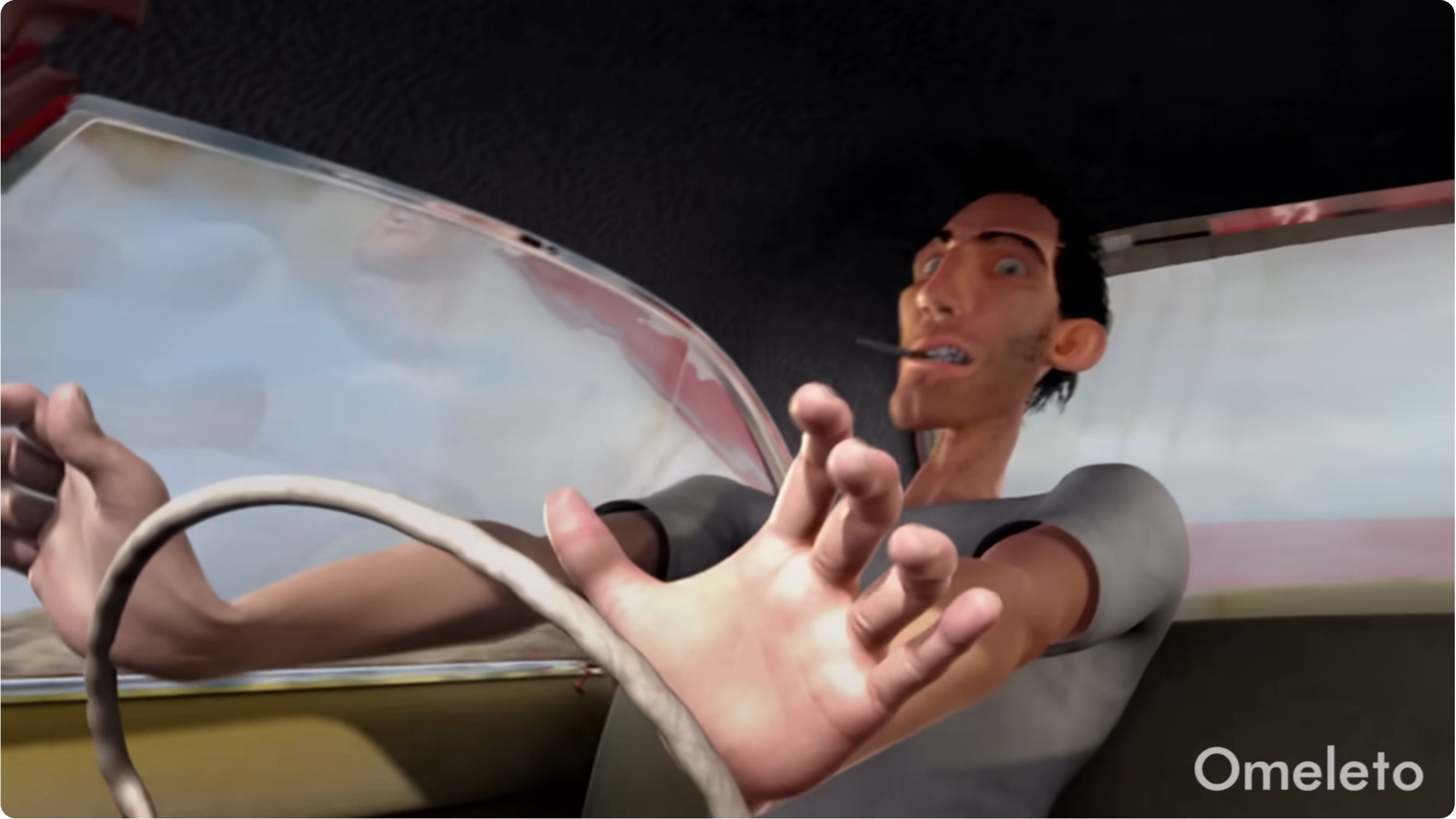} &
          \includegraphics[width=\linewidth]{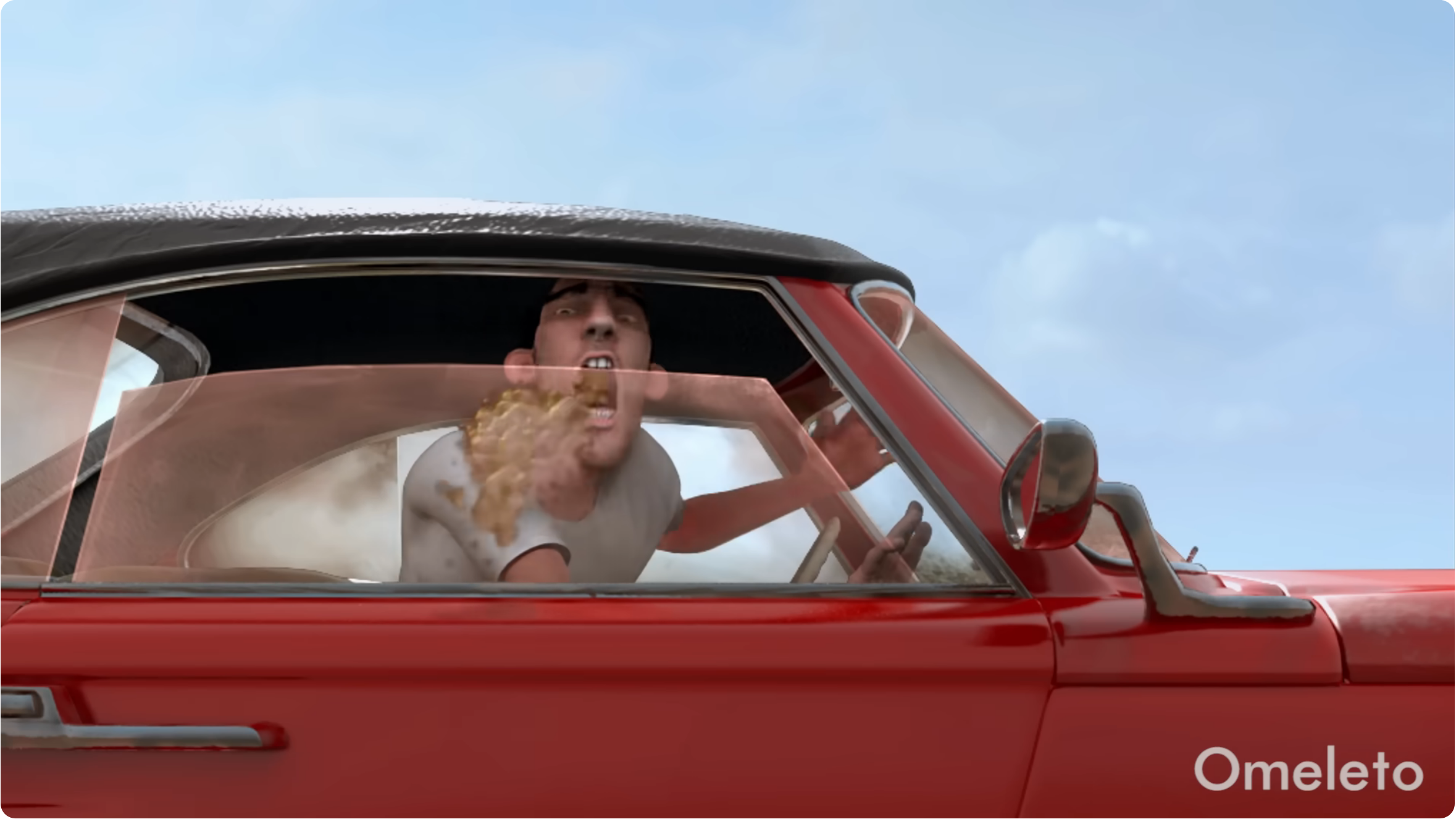} \\
        \midrule
        \textbf{GT} & The man looks for cigarettes. &
        \textcolor{darkgreen}{The man suddenly brakes to avoid a car accident.} &
        The man vomits inside his car. \\
        \cmidrule(lr){2-4}
        \textbf{Pred} & The man looks for cigarettes. &
        \textcolor{firebrick}{The man finds cigarettes and lights one but suddenly the tire of his car bursts.} &
        The man vomits inside his car. \\
    \end{tabularx}

    \caption{\textbf{Qualitative analysis.} Comparison of ground-truth chains (top) vs. GPT-4.1-mini predictions (bottom). The examples illustrate successful physical (Row 1) and motivational reasoning (Row 2) contrasted with failures due to missed fine-grained visual details (Row 3) and hallucinations (Row 4).}
    \label{fig:quali_suppl}
\end{figure*}

We present a qualitative analysis of the baselines performance using examples from GPT-4.1-mini, as illustrated in Figure~\ref{fig:quali_suppl}. These examples highlight both the emerging reasoning capabilities of VLMs and their persistent limitations in tracking specific causal mechanisms.

\noindent\textbf{Success cases.}
In the first example (Row 1), the model successfully identifies the use of the stretcher as a physical solution for crossing the gap.
In the second example (Row 2), the model understands the key reason, the vending machine getting stuck, that drives the man to hit the vending machine.

\noindent\textbf{Failure cases.} Despite these successes, the model frequently struggles with complex visual causalities, often resorting to generic events or hallucinations.
In the third example (Row 3), the model does not notice the brief action that leads to the helicopter falling from the mountain (\textit{``The rescue worker hits the helicopter's propellers with his skis''}); instead, it generates a generic event (\textit{``The helicopter takes off and approaches the mountain''}), making the subsequent crash logically inexplicable.
In the fourth example (Row 4), the model fails to connect the inattentiveness of the driver looking for cigarettes and near collision with another car, causing the driver to suddenly and violently brake, making him sick and causing him to vomit inside his car. Instead, it hallucinates a completely unrelated event: \textit{``The man finds cigarettes and lights one but suddenly the tire of his car bursts''}. This illustrates a tendency for models to generate events that are logically plausible but not grounded in the video content.

\section{Prompts}
\label{supmat:prompts}

In this section, we provide the exact prompts used for both the automated chain-construction pipeline (Section~\ref{subsec:benchmark_creation}) and the evaluation of Vision-Language Models (VLMs) (Section~\ref{subsec:baselines}).

\begin{figure*}[t]
    \centering
    \includegraphics[width=\linewidth, height=\textheight, keepaspectratio]{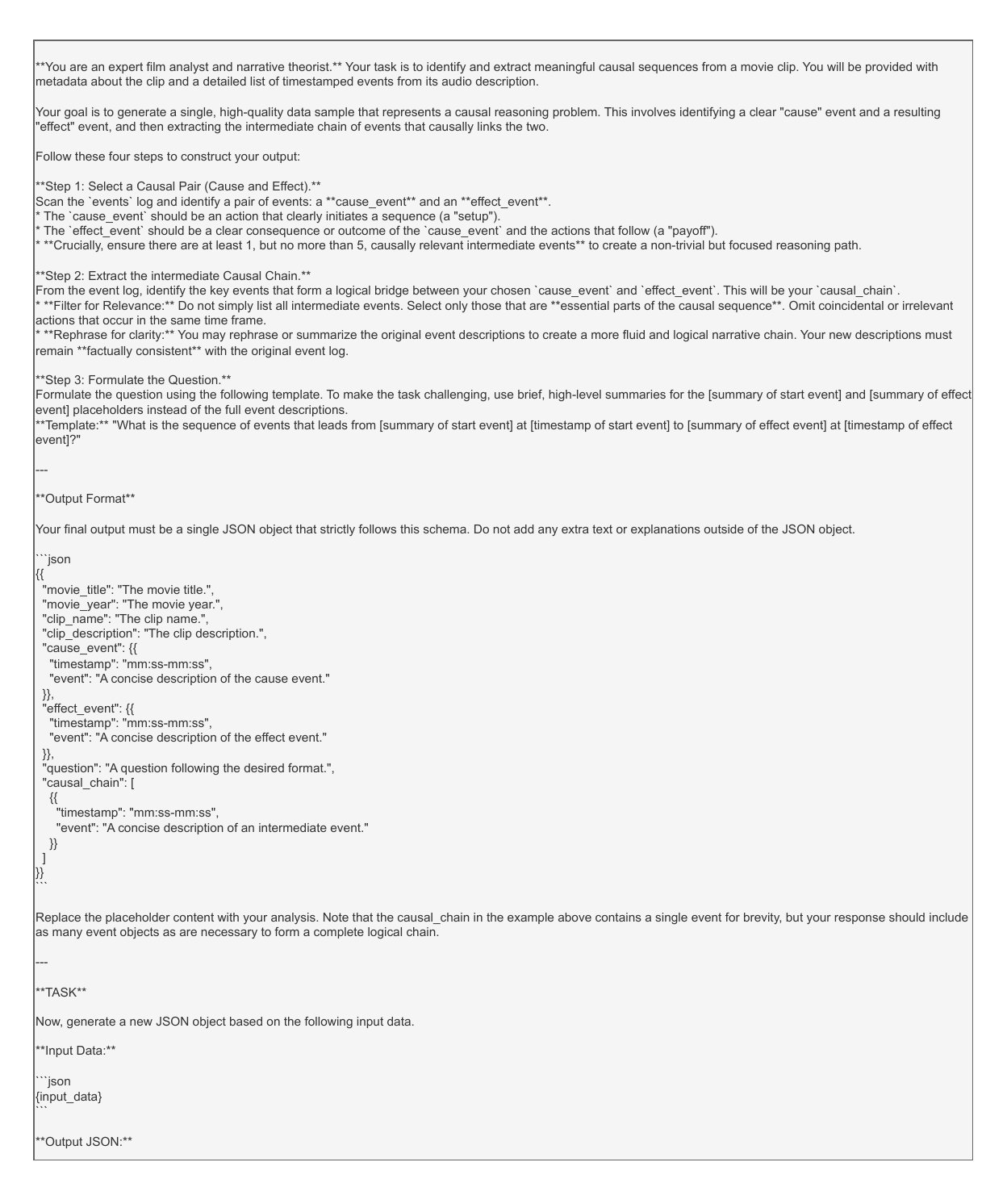}
    \caption{\textbf{Synthetic data generation prompt.} The prompt used to instruct the LLM to filter raw Audio Descriptions and extract structured causal chains for the training set generation pipeline.}
    \label{supmat:prompt_data_gen}
\end{figure*}

\noindent\textbf{Synthetic data generation.}
To transform the raw Audio Descriptions (AD) from the CMD-AD dataset into structured causal chains, we employ a Large Language Model (LLM) to filter and link events. We provided the LLM with the instructions detailed in Figure~\ref{supmat:prompt_data_gen}.

\begin{figure*}[t]
    \centering
    \includegraphics[width=\linewidth, height=\textheight, keepaspectratio]{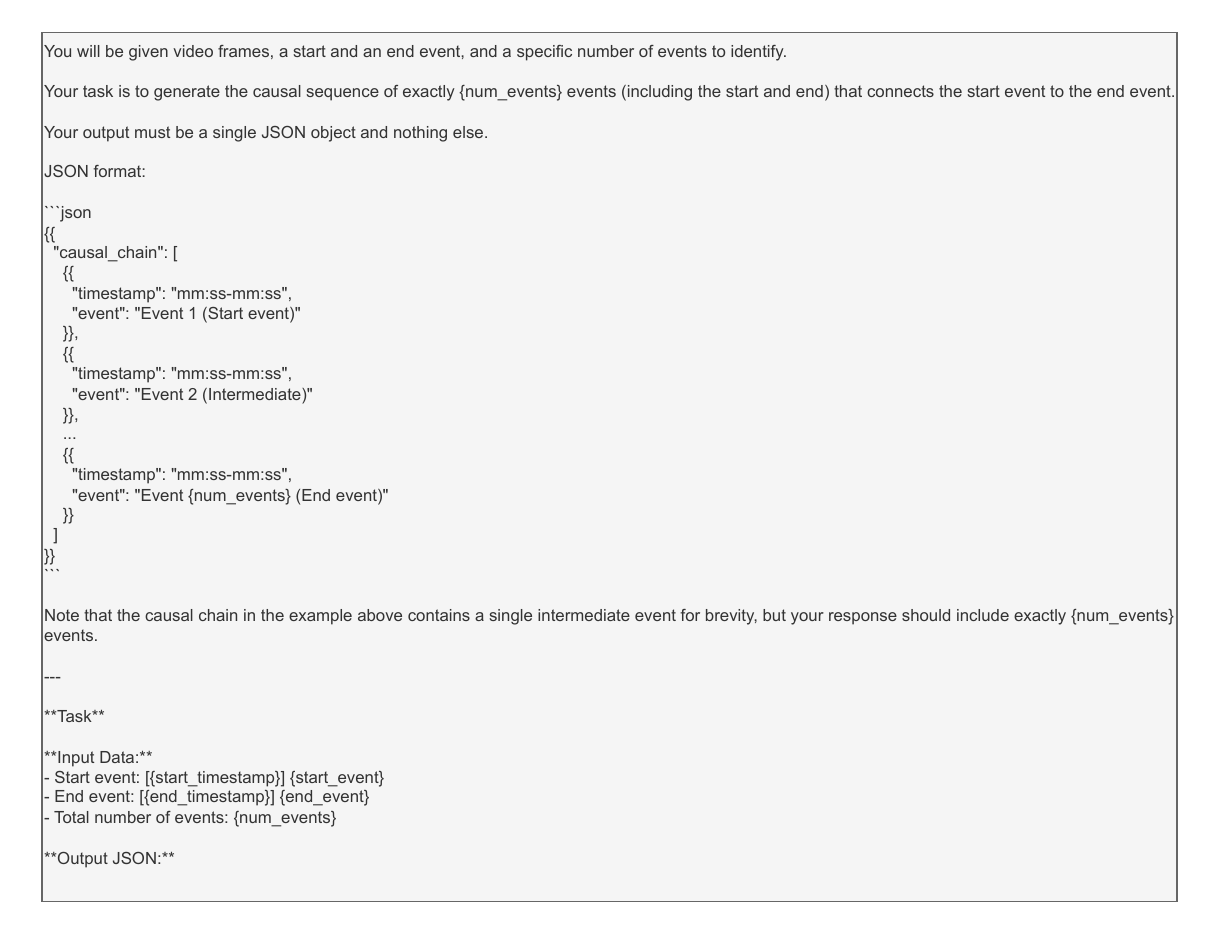}
    \caption{\textbf{Evaluation prompt.} The unified prompt used to evaluate all baseline VLMs (GPT-4.1, Gemini-2.5, Qwen3-VL) on the Causal Chain Prediction task.}
    \label{supmat:prompt_baselines}
\end{figure*}

\noindent\textbf{Baselines prompt.}
For the evaluation of all baseline models (GPT-4.1, Gemini-2.5, Qwen3-VL), we use a unified prompt structure to ensure a fair comparison (see Figure~\ref{supmat:prompt_baselines}).

\end{document}